\documentclass[final]{IEEEtran}
\IEEEoverridecommandlockouts

\usepackage{cite}
\usepackage{amsmath,amssymb,amsfonts}
\usepackage{algorithmic}
\usepackage{graphicx}
\usepackage{textcomp}
\usepackage{xcolor}
\usepackage{comment}
\usepackage{mathtools}

\usepackage{float}
\usepackage[short]{optidef}
\usepackage{cleveref}
\usepackage{algorithm}
\usepackage{algorithmic}
\usepackage{subcaption}
\usepackage{graphicx}
\usepackage[acronym]{glossaries}
	\usepackage{amsthm}
	\theoremstyle{plain}
	\newtheorem{theorem}{Theorem}

\usepackage{xcolor, soul}
\usepackage{booktabs}
\usepackage{multirow,bigdelim}
\usepackage{blindtext}
\usepackage{clipboard}

\newcommand{\numberofusers}{K} 
\newcommand{\user}{k}
\newcommand{\userSet}{\mathcal{\numberofusers}}
\newcommand{\datasetsize}{D}
\newcommand{\dataset}{\mathcal{\datasetsize}}
\newcommand{\scheduled}{s}
\newcommand{\scheduledvec}{\boldsymbol{\scheduled}}
\newcommand{\schedulingMatrix}{\boldsymbol{S}}
\newcommand{\numberofresourceblocks}{B}
\newcommand{\resourceblock}{b}

\newcommand{\inputsvec}{\boldsymbol{x}}

\newcommand{\channel}{h}
\newcommand{\channelEst}{\hat{\channel}}
\newcommand{\resourceallocation}{\lambda}
\newcommand{\resourceallocationvec}{\boldsymbol{\lambda}}
\newcommand{\resourceallocationMat}{\boldsymbol{\Lambda}}
\newcommand{\sinr}{\gamma}
\newcommand{\sinrEst}{\hat{\sinr}}
\newcommand{\sinrTH}{\sinr_{0}}
\newcommand{\e}{\varepsilon}
\newcommand{\auxilary}{\nu}
\newcommand{\auxilaryX}{l}
\newcommand{\q}{q}
\newcommand{\qX}{g}
\newcommand{\trade}{\phi}
\newcommand{\coeffKnowledge}{\varphi}

\newcommand{\lossfunc}{f}
\newcommand{\rate}{r}
\newcommand{\power}{p}
\newcommand{\noise}{N_0}
\newcommand{\interference}{I}
\newcommand{\Dual}{\psi}

\newcommand{\re}{\varrho}
\newcommand{\dualvar}{\theta}
\DeclareMathOperator*{\argmax}{argmax}
\DeclareMathOperator*{\argmin}{argmin}
\newcommand{\model}{w}
\newcommand{\modelvec}{\boldsymbol{\model}}
\newcommand{\knowledge}{j}
\newcommand{\knowledgevec}{\boldsymbol{\knowledge}}
\newcommand{\gpr}{J}
\newcommand{\covariance}{c}
\newcommand{\covarianceVec}{\covariance}
\newcommand{\covarianceMat}{\boldsymbol{C}}
\newcommand{\gprLength}{\zeta_1}
\newcommand{\gprPeriod}{\zeta_2}

\newcommand{\trainingDuration}{T}

\newcommand{\transpose}{^\dagger}
\newcommand{\optimal}{^\star}
\newcommand{\one}{\mathbf{1}}
\newcommand{\zero}{\mathbf{0}}
\newcommand{\computationtime}{\tau}
\newcommand{\clockfreq}{\omega}
\newcommand{\constantclockcyclesperdatasample}{\mu_\text{c}}
\newcommand{\voltageofcpu}{V}
\newcommand{\powerconsumptionofcomputation}{P_\text{c}}
\newcommand{\diricheletparameter}{\alpha}
\newcommand{\constantfreqpower}{b}
\newcommand{\numberoflocaliterations}{M}
\newcommand{\thetaw}{\Theta}
\newcommand{\alphaw}{\Omega}
\newcommand{\aw}{\Phi}

\DeclareMathOperator*{\sumsum}{\sum\sum}

\newcommand{\vanilaFL}{\text{IDEAL}}
\newcommand{\propPerfect}{\text{QAW}}
\newcommand{\propImperfect}{\text{QAW-GPR}}
\newcommand{\basePaper}{\text{QUNAW}}
\newcommand{\baseRand}{\text{RANDOM}}
\newcommand{\basePF}{\text{PF}}

\usepackage{soul}
\usepackage{xcolor}
\def\showcomments{1}
\newcommand{\sps}[2]{%
    \if\showcomments#1%
    \textcolor{red}{[#2]}%
    \else
    \textcolor{red}{[\st{#2}]}%
    \fi}

\newacronym{rb}{RB}{resource block}

\newacronym{fl}{FL}{federated learning}

\newacronym{Fl}{FL}{Federated learning}

\newacronym{csi}{CSI}{channel state information}

\newacronym{sgd}{SGD}{Stochastic gradient decent}

\newacronym{gpr}{GPR}{Gaussian process regression}

\newacronym{IID}{IID}{independent and identically distributed}

\newacronym{cnn}{CNN}{convolutional neural network}

\newacronym{nn}{NN}{neural network}

\newacronym{bs}{BS}{base station}

\newacronym{gp}{GP}{Gaussian process}

\newacronym{cqi}{CQI}{channel quality index}

\newacronym{lte}{LTE}{long term evolution}

\newacronym{ps}{PS}{parameter server}

\newacronym{ml}{ML}{machine learning}

\newacronym{psi}{CPSI}{computing power state information}

\newacronym{msi}{MSI}{model state information}

\newacronym{mimo}{MIMO}{multiple-input and multiple-output}

\newacronym{lp}{LP}{linear program}

\newacronym{dpp}{DPP}{dual-plus-penalty}

\newacronym{ipm}{IPM}{interior point method}

\newacronym{sinr}{SINR}{signal to interference plus noise ratio}

\begin{document}

\title{Joint Client Scheduling and Resource Allocation under Channel Uncertainty in Federated Learning}
\author{\IEEEauthorblockN{Madhusanka Manimel Wadu, 
	Sumudu Samarakoon, 
	and Mehdi Bennis}\\
\IEEEauthorblockA{\textit{Centre for Wireless	Communications (CWC), University of Oulu,
Finland} \\
\{madhusanka.manimelwadu,~sumudu.samarakoon,~mehdi.bennis\}@oulu.fi 
}

\thanks{ This work is supported by Academy of Finland 6G Flagship (grant no. 318927) and project SMARTER, projects EU-ICT IntellIoT and EUCHISTERA LearningEdge, Infotech-NOOR. A preliminary version of this work appears in the proceedings of IEEE WCNC 2020 \cite{wadu2020federated} and first author's thesis work \cite{wadu2019communication}}
}
\maketitle
\begin{abstract}
	The performance of federated learning (FL) over wireless networks depend on the reliability of the client-server connectivity and clients’ local computation capabilities. In this article we investigate the problem of client scheduling and resource block (RB) allocation to enhance the performance of model training using FL, over a pre-defined training duration under imperfect channel state information (CSI) and limited local computing resources. First, we analytically derive the gap between the training losses of FL with clients scheduling and a centralized training method for a given training duration. Then, we formulate the gap of the training loss minimization over client scheduling and RB allocation as a stochastic optimization problem and solve it using Lyapunov optimization. A Gaussian process regression-based channel prediction method is leveraged to learn and track the wireless channel, in which, the clients’ CSI predictions and computing power are incorporated into the scheduling decision. Using an extensive set of simulations, we validate the robustness of the proposed method under both perfect and imperfect CSI over an array of diverse data distributions. Results show that the proposed method reduces the gap of the training accuracy loss by up to $ 40.7\,\%$ compared to state-of-the-art client scheduling and RB allocation methods.
\end{abstract}


\glsresetall 

\section{Introduction}
\label{int}

The staggering growth of data generated at the edge of wireless networks sparked a huge interest in \gls{ml} at the network edge, coined \emph{edge \gls{ml}} \cite{park2018wireless}. In  edge \gls{ml}, training data is unevenly distributed over a large number of devices,  and  every device has a tiny fraction  of  the  data. One of the most popular model training methods in edge \gls{ml} is \emph{\gls{fl}} \cite{yang2020federated, kairouz2019advances}. The goal of \gls{fl} is to train a high-quality centralized model in a decentralized manner, based on local model training and client-server communication while training data remains private \cite{yang2020federated,konevcny2016federated,park2018wireless}. 
Recently, \gls{fl} over wireless networks gained much attention 
in communication and networking with applications  spanning a wide range of domains and verticals ranging from vehicular communication to ,blockchain and healthcare \cite{konevcny2016federated, kairouz2019advances, kim2019blockchained}.
The performance of \gls{fl} highly depends on the communication and link connectivity in addition to the model size and the number of clients engaged in training.
The quality of the trained model (inference accuracy) also depends on the training data distributions over devices which generally is non-\gls{IID} \cite{gornitz2014learning, balcazar1997computational}. Hence, the impact of training data distribution in terms of balanced-unbalancedness and \gls{IID} versus non-\gls{IID}ness  on model training is analyzed in few works \cite{kairouz2019advances, hsu2019measuring, zhao2018federated}. In \cite{hsu2019measuring}, the authors have analyzed the effect of non-\gls{IID} data distribution through numerical simulations for a visual classification task. Likewise in \cite{zhao2018federated}, authors have empirically analyzed the impact of non-\gls{IID} data distribution on the performance of \gls{fl}.

Except a handful of works \cite{chen2018lag,nishio2019client,yang2019scheduling,chen2019joint,yang2020federated,sun2020energy,amiri2020update}, the vast majority of the existing literature assumes ideal client-server communication conditions, overlooking channel dynamics and uncertainties.
In \cite{ chen2018lag}, communication overhead is reduced by using the \emph{lazily aggregate gradients (LAG)} based on reusing outdated gradient updates.
Due to the limitations in communication resources, scheduling the most informative clients is one possible solution \cite{nishio2019client, yang2019scheduling, amiri2020update}. In \cite{nishio2019client} authors propose a client-scheduling algorithm for \gls{fl} to maximize the number of scheduled clients assuming that communication and computation delays are less than a predefined threshold but, the impact of client scheduling was not studied.
In \cite{yang2019scheduling}, the authors study the impact of conventional scheduling policies (e.g., random, round robin, and proportional fairness) on the accuracy of \gls{fl} over wireless networks relying on known channel statistics.
In \cite{chen2019joint}, the training loss of \gls{fl} is minimized by joint power allocation and client scheduling.
A probabilistic scheduling framework is proposed in \cite{ren2020scheduling}, seeking the optimal trade-off between channel quality and importance of model update considering the impact of channel uncertainty in scheduling.
Moreover, the impact of dynamic channel conditions on \gls{fl} is analyzed only in few works \cite{yang2019scheduling, yang2020federated, amiri2020federated}. In \cite{yang2020federated}, authors propose over-the-air computation-based approach leveraging the ideas of \cite{4305404} to speed-up the global model aggregation utilizing the superposition property of a wireless multiple-access channels with scheduling and beamforming.
In addition to channel prediction, clients' computing power are utilized for their local models updates.
\gls{sgd} is widely used for updating their local models, in which the computation is executed sample by sample \cite{bottou2010large}.
In \cite{amiri2020federated}, the authors proposed the compressed analog distributed stochastic gradient descent (CA-DSGD) method,  which is shown to be robust against imperfect \gls{csi} at the devices. While interestingly the communication aspects in \gls{fl} such as optimal client scheduling and resource allocation in the absence of perfect \gls{csi} along with the limitations in processing power for \gls{sgd}-based local computations are neglected in all these aforementioned works.

Acquiring \gls{csi} through pilot sequence exchanges introduces an additional signaling overhead that scales with the number of devices.
There are a handful of works dealing with wireless channel uncertainties \cite{duel2007fading, liu2006recurrent}.
Authors in \cite{duel2007fading} demonstrated the importance of reliable fading prediction for adaptive transmission in wireless communication systems.
Channel prediction via fully connected recurrent neural networks are proposed in \cite{liu2006recurrent}.
Among channel prediction methods, \gls{gpr} is a light weight online technique where the objective is to approximate a function with a non-parametric Bayesian approach under the presence of nonlinear and unknown relationships between variables \cite{karaca2016entropy, osborne2007gaussian}. A \gls{gp} is a stochastic process with a collection of random variables indexed by time or space. Any subset of these random variables forms multidimensional joint Gaussian distributions, in which \gls{gp} can be completely described by their mean and covariance matrices.
\Copy{motivationgaussian}{{The foundation of the \gls{gpr}-approach is Bayesian inference, where a priori model is chosen and updated with observed experimental data \cite{karaca2016entropy}.
In the literature, \gls{gpr} is used for a wide array of practical applications including communication systems \cite{perez2013gaussian,schwaighofer2003gpps,chiumento2015adaptive}.
In \cite{perez2013gaussian} \gls{gpr} is used to estimate Rayleigh channel.
In \cite{schwaighofer2003gpps}, a problem of  localization in a cellular network is investigated with \gls{gpr}-based possition predictions.
In \cite{chiumento2015adaptive}, \gls{gpr} is used to predict the \gls{cqi} to reduce \gls{cqi} signaling overhead.}}
In our prior works, \cite{wadu2020federated, wadu2019communication}, \gls{gpr}-based channel prediction is used to derive a client scheduling and \gls{rb} allocation policy under imperfect channel conditions assuming unlimited computational power availability per client. Investigating the performance of \gls{fl} under clients' limited computation power under both \gls{IID} and non-\gls{IID} data distributions remains an unsolved problem.

The main contributions of this work over \cite{wadu2020federated, wadu2019communication} are the derivation of a joint client scheduling and \gls{rb} allocation policy for \gls{fl} under communication and computation power limitations and a comprehensive analysis of the performance of model training as a function of (i) system model parameters, (ii) available computation and communication resources, (iii) non-\gls{IID}  data distribution over the clients in terms of the heterogeneity of dataset sizes and available classes.
In this work, we consider a set of clients  that communicate with a server over wireless links to train a \gls{nn} model within a predefined training duration.
First, we derive an analytical expression for the loss of accuracy in FL with scheduling compared to a centralized training method.
In order to reduce the signaling overhead in pilot transmission for channel estimation, we consider the communication scenario with imperfect \gls{csi} and we leverage \gls{gpr} to learn and track the wireless channel while quantifying the information on the unexplored \gls{csi} over the network.
To do so, we formulate the client scheduling and \gls{rb} allocation problem as a trade-off between optimal client scheduling, \gls{rb} allocation, and \gls{csi} exploration under both communication and computation resource  constraints.
\Copy{noveltypage}{{Due to the stochastic nature of the aforementioned problem, we resort to the \gls{dpp} technique from the Lyapunov optimization framework to recast the problem into a set of linear problems that are solved at each time slot \cite{neely2010stochastic}.
In this view, we present the joint client scheduling and RB allocation algorithm that simultaneously explore and predict \gls{csi} to improve the accuracy of \gls{fl} over wireless links.}}
With an extensive set of simulations we validate the proposed methods over an array of \gls{IID} and non-\gls{IID} data distributions capturing the heterogeneity in dataset sizes and available classes. In addition, we compare the feasibility of the proposed methods in terms of fairness of the trained global model and accuracy.
Simulation results show that the proposed method achieves up to $ 40.7\,\%$ reduction in the loss of accuracy compared to the state-of-the-art client scheduling and RB allocation methods.

The rest of this paper is structured as follows.
Section \ref{pr} presents the system model and formulates the problem of model training over wireless links under imperfect \gls{csi}.
In Section \ref{sec:solution}, the problem is first recast in terms of the loss of accuracy due to client scheduling compared to a centralized training method. Subsequently, \gls{gpr}-based \gls{csi} prediction is proposed followed by the derivation of Lyapunov optimization-based client scheduling and \gls{rb} allocation policies under both perfect and imperfect \gls{csi}. Section \ref{res} numerically evaluates the proposed scheduling policies over state-of-the-art client scheduling techniques. Finally, conclusions are drawn in Section \ref{conclu}.

\section{System Model and Problem Formulation}
\label{pr}
Consider a system consisting of a set $ \mathcal{\numberofusers} $ of $ \numberofusers $ clients that communicate with a \gls{ps} over wireless links.
Therein, the $ \user $-th client has a private dataset $ \mathcal{\dataset}_\user $ of size $\datasetsize_{\user} $, which is a partition of the global dataset $ \mathcal{\dataset} $ of size
$ \datasetsize = \sum_{\user} \datasetsize_{\user} $. For communication with the server for local model updating, a set $ \mathcal{\numberofresourceblocks} $ of $ \numberofresourceblocks (\le \numberofusers $) \glspl{rb} is shared among those clients{\footnote{\Copy{footnoteresourceblock}{For $ B>K $, all clients simultaneously communicate their local models to the PS.}}}.
\subsection{Scheduling, resource block allocation, and channel estimation}

Let $ \scheduled_{\user}(t) \in \{0,1\} $ be an indicator where $\scheduled_{\user}(t) =1$ indicates that the client $ \user $ is scheduled by the \gls{ps} for uplink communication at time $ t $ and $ \scheduled_{\user}(t) =0 $, otherwise.
Multiple clients are simultaneously scheduled by allocating each at most with one \gls{rb}.
Hence, we define the RB allocation vector $\boldsymbol{\resourceallocation}_{\user}(t) = [\resourceallocation_{\user,\resourceblock}(t)]_{\forall\resourceblock\in\mathcal{\numberofresourceblocks}}$ for client $\user$ with $ \resourceallocation_{\user,\resourceblock}(t) =1 $ when RB $ \resourceblock $ is allocated to client $\user$ at time $t$, and $ \resourceallocation_{\user,\resourceblock}(t) = 0 $, otherwise.
The client scheduling and RB allocation are constrained as follows:
\begin{equation}\label{eq:scheduleresource}
    \scheduled_{\user}(t) 
    \leq
    \one\transpose \boldsymbol{\resourceallocation}_{\user}(t)
    \leq 
    1
    \quad \forall \user, t,
\end{equation}
where $\one \transpose$ is the transpose of the all-one vector.
The rate at which the $k$-th client communicates in the uplink with the \gls{ps} at time $ t $ is given by,
\begin{equation}\label{rate}
\rate_{\user}(t)  
= 
\textstyle \sum_{\resourceblock \in \mathcal{\numberofresourceblocks}}
\resourceallocation_{\user,\resourceblock}(t)
\log_2 \big( 1 + 
\frac{\power |\channel_{\user,\resourceblock}(t)|^2} { \interference_{\user,\resourceblock}(t) + \noise }
\big),
\end{equation}
where $\power$ is a fixed transmit power of client $ \user $, $ \channel_{\user,\resourceblock} (t) $ is the channel between client $ \user$ and the \gls{ps} over \gls{rb}  $ \resourceblock$ at time $ t $, $\interference_{\user,\resourceblock}(t) $ represents the uplink interference on client $\user$ imposed by other clients over \gls{rb} $\resourceblock$, and $ \noise $ is the noise power spectral density.
{Note that, a successful communication between a scheduled client and the server within a channel coherence time is defined by satisfying a target minimum rate.}
In this regard, a rate constraint is imposed on the \gls{rb} allocation in terms of a target signal to interference plus noise ratio (SINR) $\sinrTH$ as follows:
\begin{equation}\label{eq:allocationindicator}
\resourceallocation_{\user,\resourceblock}(t) 
\leq 
\mathbb{I} \big( \sinrEst_{\user,\resourceblock}(t) \geq \sinrTH \big)
\quad \forall \user,\resourceblock,t,
\end{equation}
where
$\sinrEst_{\user,\resourceblock}(t) = \frac{\power |\channel_{\user,\resourceblock}(t)|^2} { \interference_{\user,\resourceblock}(t) + \noise }$ 
and the indicator $\mathbb{I}(\sinrEst \geq \sinrTH) =1$ only if $\sinrEst \geq \sinrTH$.

\subsection{Computational power consumption and computational time}
Most of the clients in wireless systems are mobile devices operated by power-limited energy sources, with limited computational capabilities.
As demonstrated in \cite{yuan2006energy} the dynamic power consumption $ \powerconsumptionofcomputation $ of a processor with clock frequency of $\clockfreq$ is proportional to the product of $ \voltageofcpu ^2 \clockfreq $.
Here, $ \voltageofcpu $ is the supply voltage of the computing, which is approximately linearly proportional to $\clockfreq$ \cite{de2013energy}.
Motivated by \cite{yuan2006energy} and \cite{de2013energy}, in this work, we adopt the model $ \powerconsumptionofcomputation \propto \clockfreq^3 $ to capture the computational power consumption per client.
Due to the concurrent tasks handled by the clients in addition to local model training, the dynamics of the available computing power $ \powerconsumptionofcomputation $ at a client is modeled as a Poison arrival process \cite{castro2019modeling}. Therefore, we assume that the computing power $ \powerconsumptionofcomputation^\user(t) $ of client $\user$ at time $t$ follows an independent and identically distributed exponential distribution with mean $\constantclockcyclesperdatasample$. As a result, the minimum computational time required for client $ \user $ is given by,
\begin{equation}\label{computationtimeandclockfrequency}
	\computationtime_{\text{c},{\user}}(t) = \frac{\constantclockcyclesperdatasample \datasetsize_{\user}\numberoflocaliterations_{\user}}{\sqrt[3]{\powerconsumptionofcomputation^\user(t)} / \constantfreqpower},
\end{equation}
where $ \numberoflocaliterations_{\user}  $ is the number of \gls{sgd} iterations computed over $ \datasetsize_{\user} $. The constants $ \constantclockcyclesperdatasample $ and $ \constantfreqpower $ are the number of clock cycles required to process a single sample and power utilized per number of computation cycles, respectively.
{To avoid unnecessary delays in the overall training that are caused by clients' local computations}, we impose a constraint on the computational time, with threshold $ \computationtime_{0} $ over the scheduled clients as follows:

\begin{equation}\label{computationaltime}
	\scheduled_{\user}(t) \leq \mathbb{I}(\computationtime_{\text{min},{\user}}(t) \leq \computationtime_{0})\quad \forall \user,t.
\end{equation}

\subsection{Model training using \gls{fl}}

The goal in \gls{fl} is to minimize a regularized loss function $F(\modelvec,\mathcal{\dataset}) = \frac{1}{\datasetsize}
\sum_{\boldsymbol{\inputsvec}_i\in\mathcal{\dataset}}\lossfunc(\boldsymbol{\inputsvec}_i\transpose\modelvec ) + \xi \re( \modelvec)$, which is parametrized by a weight vector $\modelvec$ (referred as the \emph{model}) over the global dataset within a predefined communication duration $\trainingDuration$.
Here, $\lossfunc(\cdot)$ captures a loss of either regression or classification task, $ \re (\cdot) $ is a regularization function such as Tikhonov, $ \boldsymbol{\inputsvec}_i $ is the input vector, and $\xi (>0)$ is the regularization coefficient.
It is assumed that, clients do not share their data and instead, compute their models over the local datasets.

For distributed model training in \gls{fl}, clients share their local models that are computed by solving $ F(\modelvec,\mathcal{\dataset}) $ over their local datasets using \gls{sgd} using the \gls{ps}, which in return does model averaging and shares the global model with all the clients. 

Under imperfect \gls{csi}, prior to transmission the channels need to be estimated via sampling.
Instead of performing channel measurements beforehand the inferred channels between clients and the \gls{ps} over each \gls{rb} are as $\channelEst(t) = \gpr\big(t, \{\tau,\channel(\tau)\}_{\tau\in\mathcal{N}(t)} \big)$ using past $ N $ channel observations.
\Copy{channelpredictclaim}{{Under channel prediction, $ \channel_{\user,\resourceblock} (t) = \hat{\channel}_{\user,\resourceblock} (t) $ is used as \gls{sinr} in \eqref{rate} and \eqref{eq:allocationindicator}.}}
The choice of small $ N $ ensures low computation complexity of the inference task.
Here, the set $\mathcal{N}(t)$ consists of $ N $ most recent time instances satisfying $\scheduled(\tau)=1$ and $\tau<t$.
With $\resourceallocation_{\user,\resourceblock}(t)=1$, the channel $\channel_{\user,\resourceblock}(t)$ is sampled and used as an observation for the future, i.e., the \gls{rb} allocation and channel sampling are carried out simultaneously.
In this regard, we define the information on the channel between client $ \user $ and the server at time $t$ as $\knowledgevec_\user(t) = [\knowledge_{\user,\resourceblock}(t)]_{\resourceblock \in \mathcal{\numberofresourceblocks}}$.
For accurate CSI predictions, it is essential to acquire as much information about the CSI over the network \cite{karaca2012smart}.
This is done by exploring new information by maximizing $\sum_{\user} \knowledgevec_\user\transpose(t) \resourceallocationvec_\user(t)$ at each time $t$ while minimizing the loss $F(\modelvec,\mathcal{\dataset})$.
\begin{figure}[!t]
	\centering
	\includegraphics[width = \columnwidth]{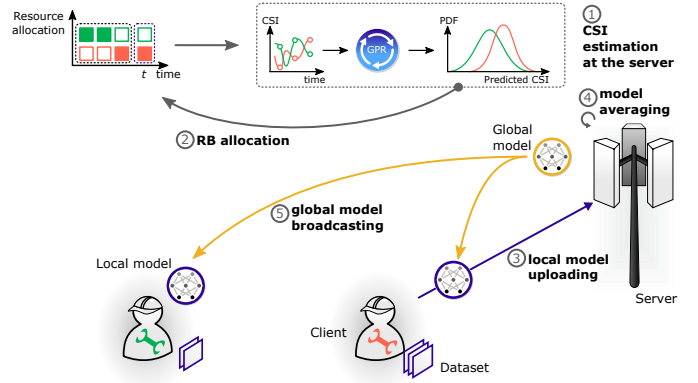}
	\caption{FL with client scheduling under limited wireless resources and imperfect CSI.}
	\label{fig:system_model}
\end{figure}
Therefore, the empirical loss minimization problem over the training duration is formally defined as follows:
\begin{subequations}\label{eq:master_problem}
\begin{align}
\label{obj:master_problem}
\underset{\modelvec(t), \scheduledvec(t), \resourceallocationMat(t), \forall t}{\text{minimize}} 
& F\big( \modelvec(T), \mathcal{\datasetsize} \big) 
- 
\textstyle 
\frac{\coeffKnowledge}{\trainingDuration}
\sum_{\user,t} \knowledgevec_\user\transpose(t) \resourceallocationvec_\user(t),
\\
\label{cns:model_constraints}
\text{subject to\hphantom{00}} 
& \eqref{eq:scheduleresource}\text{-}\eqref{eq:allocationindicator},\eqref{computationaltime}, \\
\label{cns:OFDMA}
& \boldsymbol{A}\resourceallocationMat\transpose(t) \preceq \one \quad \forall t,\\
\label{cns:RB_availability}
& \one\transpose\scheduledvec(t) \leq \numberofresourceblocks \quad \forall t,\\
\label{cns:boolean}
& \scheduledvec(t) \in \{0,1\}^\numberofusers,\resourceallocationvec_\user(t) \in \{0,1\}^\resourceblock \quad \forall t,\\
\label{cns:local_SGD}
& \modelvec_{\user}(t) = { \argmin_{\modelvec'} F( \modelvec'| \modelvec(t-1),\dataset_\user}) \quad \forall k, t \\
\label{cns:model_update}
& \textstyle \modelvec(t) = \sum_\user \frac{\datasetsize_\user}{\datasetsize} \scheduled_\user(t) \modelvec_\user(t) \quad \forall t, 
\end{align}
\end{subequations}
where 
$\resourceallocationMat\transpose(t) = [\resourceallocationvec_\user\transpose(t)]_{\user\in\userSet}$,
$\coeffKnowledge(>0)$ controls the impact of the \gls{csi} exploration,
and
$\boldsymbol{A}$ is a $\numberofresourceblocks\times\numberofusers$ all-one matrix.
The orthogonal channel allocation in \eqref{cns:OFDMA} ensures collision-free client uplink transmission with $I_{\user,\resourceblock}(t)=0$ and constraint \eqref{cns:RB_availability} defines the maximum allowable clients to be scheduled.
The \gls{sgd} based local model calculation at client $\user$ is defined in \eqref{cns:local_SGD}.
\Copy{gammathreshold}{{The choice of \gls{sinr} target $ \sinrTH $ in \eqref{eq:allocationindicator} ensures that local models are uploaded within a single coherence time interval}}{\footnote{\Copy{gammathresholdfootnote}{The prior knowledge on channel statics, model size, transmit power, and bandwidth is used to choose $ \sinrTH $.}}.
}

\section{Optimal client-Scheduling and RB Allocation Policy via Lyapunov Optimization}\label{sec:solution}
The optimization problem \eqref{eq:master_problem} is coupled over all clients hence, in what follows, we elaborate on the decoupling of \eqref{eq:master_problem} over clients and the server, and derive the optimal client scheduling and RB allocation policy.

\subsection{Decoupling loss function of \eqref{eq:master_problem} via dual formulation}
\label{dualformulationsection}
Let us consider an ideal unconstrained scenario where the server gathers all the data samples and trains a global model in a \emph{centralized} manner.
Let $F_0 = \min_{\modelvec} F(\modelvec,\dataset)$ be the minimum loss under centralized training.
By the end of the training duration $\trainingDuration$, we define the gap between the proposed \gls{fl} under communication constraints and centralized training as $\e(\trainingDuration) = F\big( \modelvec(T), \mathcal{\datasetsize} \big) - F_0$.
In other words, $\e(\trainingDuration)$ is the accuracy loss of \gls{fl} with scheduling compared to centralized training.
Since $ F_0 $ is independent of the optimization variables in \eqref{obj:master_problem}, replacing $ F(\modelvec,\dataset) $ by $\e(\trainingDuration)$ does not affect optimality under the same set of constraints.

\Copy{theequation}{{To analyse the loss of \gls{fl} with scheduling, we consider the dual function of \eqref{obj:master_problem} with the dual variable $\boldsymbol{\dualvar}=[\dualvar_1,\ldots,\dualvar_{\datasetsize}]$, $\boldsymbol{X}=[\boldsymbol{X}_\user]_{\user\in\userSet}$ with $\boldsymbol{X}_\user = [ \inputsvec_i ]_{i=1}^{\datasetsize_\user}$, and $ \boldsymbol{z}=\boldsymbol{X}^T\boldsymbol{w}$ as follows:
\begin{subequations}
	\label{eq:conjugates_big}
	\begin{align}
	\Dual(\boldsymbol{\dualvar}) &=
	\min_{\boldsymbol{w}, \boldsymbol{z}}
	\Big(  
	\textstyle \sum_{\boldsymbol{\inputsvec}_i\in\mathcal{\datasetsize}}
	\frac{1}{\datasetsize}\lossfunc_i( \boldsymbol{\inputsvec}_i^T\boldsymbol{w} ) 
	+ \xi \re ( \boldsymbol{w})
	+ \frac{ \boldsymbol{\dualvar}^T(\boldsymbol{z}-\boldsymbol{X}^T\boldsymbol{w}) }{\datasetsize}  \Big)\\	
	&\textstyle \stackrel{a}{=}\frac{1}{\datasetsize} \underset{\boldsymbol{w}}{\inf}\big\{ \datasetsize \xi  \re ( \boldsymbol{w})-\boldsymbol{\dualvar}^T \boldsymbol{X}^T\boldsymbol{w}  \big\} \nonumber \\ &\qquad \qquad + \textstyle \sum_{i=1}^{\datasetsize} \underset{z_i}{\inf} \big\{ -\dualvar_i z_i - \lossfunc_i (z_i)\big\}\\
	&\stackrel{b}{=}\xi\underset{\boldsymbol{w}}{\sup}\big\{ \boldsymbol{w}^T \boldsymbol{v} - \re ( \boldsymbol{w}) \big\} + \nonumber \\ &\qquad \qquad \textstyle  \sum_{k=1}^{\numberofusers}  \sum_{i=1}^{\datasetsize_\user} \underset{z_i}{\sup} \big\{ \lossfunc_i (z_i) + \dualvar_i z_i\big\}\\
	&\textstyle\stackrel{c}{=} 
	-\underbrace{\textstyle  \sum_{k=1}^{\numberofusers}  \sum_{i=1}^{\datasetsize_\user} \frac{1}{\datasetsize} \lossfunc_i^* (-\dualvar_i)}_{\#1}-
	\underbrace{\xi \re^*(\boldsymbol{v})}_{\#2}.
	\label{eq:conjugates}
	\end{align}
\end{subequations}
Here, step (a) rearranges the terms, (b) converts the infimum operations to supremum while substituting $\boldsymbol{v} = {\boldsymbol{X}\boldsymbol{\dualvar}}/{\xi \datasetsize}$, and (c) uses the definition of conjugate function $ \lossfunc_i^* (-\dualvar_i) = \textstyle \underset{z_i}{\inf} \{ -\dualvar_i z_i - \lossfunc_i (z_i)\} $ and $ \re^*(\boldsymbol{v}) = \underset{\boldsymbol{w}}{\sup}\{\boldsymbol{w}^T \boldsymbol{v} -\re(\boldsymbol{w} ) \}$ \cite{boyd2004convex}.}}
With the dual formulation, the relation between the primal and dual variables is $\boldsymbol{w} = \nabla \re ^*(\boldsymbol{v})$ \cite{yang2019scheduling}.
Based on the dual formulation, the loss \gls{fl} with scheduling is $\e(\trainingDuration) = \Dual_0 - \Dual\big( \boldsymbol{\dualvar}(T) \big)$ where $\Dual_0$ is the maximum dual function value obtained from the centralized method.
Note that the first term of \eqref{eq:conjugates} decouples per client and thus, can be computed locally.
In contrast, the second term in \eqref{eq:conjugates} cannot be decoupled per client.
%
To compute $\re^*(\boldsymbol{v})$, first, each client $ \user $ locally computes $ \Delta \boldsymbol{v}_\user(t)  = \frac{1}{\xi \datasetsize} \boldsymbol{X}_{\user} \Delta \boldsymbol{\dualvar}_{\user}(t) $ at time $ t $.
Here, $ \Delta \boldsymbol{\dualvar}_k(t) $ is the change in the dual variable $ \boldsymbol{\dualvar}_k(t) $ for client $ k $ at time $ t $ given as below,
\begin{multline}\label{eq:delta_theta}
\Delta \boldsymbol{\dualvar}_{\user}(t) 
\approx
\textstyle \argmax_{ \boldsymbol{\delta} \in \mathbb{R}^{\datasetsize_\user}} 
\Big( 
-\frac{1}{\datasetsize} \one\transpose  [ \lossfunc_i^* (-\boldsymbol{\dualvar}_{\user}(t) - \boldsymbol{\delta}) ]_{i =1}^{\datasetsize_\user} 
\textstyle
\\- \frac{\xi}{\numberofusers} \re^*\big(\boldsymbol{v}(t) \big) \textstyle
-  \frac{1}{\datasetsize} \boldsymbol{\delta}\transpose \boldsymbol{X}_{k} \re^*\big(\boldsymbol{v}(t)\big) - \frac{\eta/\xi}{2 \datasetsize^2} \| \boldsymbol{X}_{\user} \boldsymbol{\delta}\|^2
\Big),
\end{multline}
where $\eta$ depends on the partitioning of $\dataset$ \cite{hiriart2004fundamentals}.
It is worth noting that $\Delta \boldsymbol{\dualvar}_k(t)$ in \eqref{eq:delta_theta} is computed based on the previous global value $\boldsymbol{v}(t)$ received by the server.
Then, the scheduled clients upload $\Delta \boldsymbol{v}_\user(t)$ to the server.
Following the dual formulation, the model aggregation and update in \eqref{cns:model_update} at the server is modified as follows:
\begin{equation}\label{model_update_new}
\boldsymbol{v}(t+1)
\coloneqq   \textstyle 
\boldsymbol{v}(t) + \sum_{\user \in \mathcal{\numberofusers}} \scheduled_{\user}(t) \Delta \boldsymbol{v}_\user(t).
\end{equation}
Using \eqref{model_update_new}, the server computes the coupled term $\re^*\big(\boldsymbol{v}(t+1)\big)$ in \eqref{eq:conjugates}.
It is worth noting that from the $t$-th update, $\Delta \boldsymbol{\dualvar}_k(t)$ in \eqref{eq:delta_theta} maximizes $\Delta\Dual\big(\boldsymbol{\dualvar}_\user(t)\big)$, which is the change in the dual function $\Dual\big(\boldsymbol{\dualvar}(t)\big)$ corresponding to client $\user$. 
Let $\boldsymbol{\dualvar}\optimal_\user(t)$ be the local optimal dual variable at time $t$, in which $\Delta\Dual\big(\boldsymbol{\dualvar}\optimal_\user(t)\big) \geq \Delta\Dual\big(\boldsymbol{\dualvar}_\user(t)\big)$ is held.
Then for a given accuracy $\beta_{\user}(t) \in (0,1) $ of local \gls{sgd} updates, the following condition is satisfied:
\begin{equation}\label{beta}
\textstyle \frac{  \Delta \Dual_\user\big(\Delta \boldsymbol{\dualvar}\optimal_{\user} (t)\big)  - \Delta \Dual_\user\big(\Delta \boldsymbol{\dualvar}_{\user} (t)\big)  }{ \Delta \Dual_\user\big(\Delta \boldsymbol{\dualvar}_{\user} (t)\big)  - \Delta \Dual_\user(0) } \leq \beta_{\user}(t),
\end{equation}
where
$ \Delta \Dual_\user(0) $ is the change in $ \Dual $ with a null update from the $ k $-th client.
For simplicity, we assume that $\beta_{\user,t} = \beta$ for all $\user\in\userSet$ and $t$, hereinafter.
With \eqref{beta}, the gap between \gls{fl} with scheduling and the centralized method is bounded as shown in Theorem \ref{theorem:gap}:

\begin{theorem}
	\label{theorem:gap}
	The upper bound of $ \e(\trainingDuration) $ after {$ T $ communication rounds} is given by,
	\[\e(\trainingDuration) \leq \datasetsize \Big( 1-(1-\beta) 
	\textstyle \sumsum_{t\leq\trainingDuration}^{\user\leq\numberofusers} \frac{\datasetsize_\user}{\trainingDuration\datasetsize} \scheduled_\user(t)
	\Big)^\trainingDuration.\]	
\end{theorem}
\begin{IEEEproof}
	See Appendix \ref{gap_proof}.
\end{IEEEproof}
This yields that the minimization of $\e(\trainingDuration)$ can be achieved by minimizing its upper bound defined in Theorem \ref{theorem:gap}.
Henceforth, the equivalent form of \eqref{eq:master_problem} is given as follows:
\begin{subequations}\label{eq:modified_problem}
	\begin{align}
	\underset{[\Delta \boldsymbol{\dualvar}_{\user}(t)]_{\user}, \scheduledvec(t), \resourceallocationMat(t), \forall t}{\text{minimize}} 
	& \datasetsize \Big( 1- (1-\beta)
	\textstyle \sum_{t,\user} \frac{\datasetsize_\user}{\trainingDuration\datasetsize} \scheduled_\user(t)
	\Big)^\trainingDuration \nonumber \\ &\qquad \qquad- \frac{\coeffKnowledge}{\trainingDuration}
	\sum_{\user,t} \knowledgevec_\user\transpose(t) \resourceallocationvec_\user(t),  \label{obj:modified_problem}\\
	\label{cns:all_modified}
	\text{subject to \hphantom{000}} 
	& \eqref{cns:model_constraints}\text{-}\eqref{cns:boolean}, \eqref{eq:delta_theta}, \eqref{model_update_new}.
	\end{align}
\end{subequations}

\subsection{\gls{gpr}-based information metric $\gpr(\cdot)$ for unexplored \gls{csi}}\label{subsec:gpr}

For \gls{csi} prediction, we use \gls{gpr} with a Gaussian kernel function to estimate the nonlinear relation of $\gpr(\cdot)$ with a \gls{gp} prior.
For a finite data set $\{t_n,\channel(t_n)\}_{n\in\mathcal{N}}$, the aforementioned \gls{gp} becomes a multi-dimensional Gaussian distribution, with zero mean and covariance $\covarianceMat = [\covariance(t_m,t_n)]_{m,n\in\mathcal{N}}$ given by,
\begin{equation}\label{eq:gpr_covar}
\covariance(t_m,t_n) = 
\textstyle \exp \Big( 
-\frac{1}{\gprLength} \sin^2 \big(\frac{\pi}{\gprPeriod} (t_m-t_n) \big)
\Big),
\end{equation}
where $\gprLength$ and $\gprPeriod$ are the length and period hyper-parameters, respectively \cite{xing2015gpr}.
Henceforth, the \gls{csi} prediction at time $t$ and its uncertainty/variance is given by \cite{PrezCruz2013GaussianPF},
\begin{eqnarray}
\label{eq:gpr_mean}
&\channelEst(t) 
= \covarianceVec\transpose(t)\covarianceMat^{-1}[\channel(t_n)]_{n\in\mathcal{N}}, \\
\label{eq:gpr_var}
&\Upsilon(t) 
= \covariance(t,t) - \covarianceVec\transpose(t)\covarianceMat^{-1}\covarianceVec(t),
\end{eqnarray}
where
$\covarianceVec(t) = [\covariance(t,t_n)]_{n\in\mathcal{N}}$.
The client and \gls{rb} dependence is omitted in the discussion above for notation simplicity.
\Copy{gprmotivation}{{
The uncertainty measure $\Upsilon(.)$ of the predicted channel $ \channelEst $ calculated using \gls{gpr} framework is used as $\knowledge(.)$ in \eqref{obj:master_problem} which in turn allowing to exploring and sampling channels with high uncertainty towards improving the prediction accuracy.
}}Finally, it is worth nothing that under \emph{perfect \gls{csi}} $\channelEst(t) = \channel(t)$ and $\knowledge(t)=0$.

\subsection{Joint client scheduling and \gls{rb} allocation}\label{subsec:policy}

Due to the time average objective in \eqref{obj:modified_problem}, the problem \eqref{eq:modified_problem} gives rise to a stochastic optimization problem defined over $t = \{1,\ldots,\trainingDuration\}$.
Therefore, we resort to the \emph{drift plus penalty} (DPP) technique from Lyapunov optimization framework to derive the optimal scheduling policy \cite{neely2010stochastic}.
Therein, the Lyapunov framework allows us to transform the original stochastic optimization problem into a series of optimizations problems that are solved at each time $t$, as discussed next.
First, we denote $u(t) = (1-\beta)\sum_\user \scheduled_\user(t)\datasetsize_\user / \datasetsize$ and define its time average $\bar{u} = \sum_{t\leq\trainingDuration} u(t) / \trainingDuration$.
Then, we introduce auxiliary variables $ \auxilary(t)$ and $\auxilaryX(t)$ with time average lower bounds $\bar{\auxilary} \leq \bar{u} $ and $\bar{\auxilaryX} \leq \frac{1}{\trainingDuration}
\sum_{\user,t} \knowledgevec_\user\transpose(t) \resourceallocationvec_\user(t) \leq \auxilaryX_0$, respectively.
To track the time average lower bounds, we introduce virtual queues $ \q(t) $ and $\qX(t)$ with the following dynamics \cite{neely2010stochastic,bethanabhotla2014adaptive, mao2015lyapunov}:

\begin{subequations}\label{eq:queue}
	\begin{eqnarray}
	&\q(t+1) = \max \big( 0 , \q(t) + \auxilary(t) - u(t) \big), \\
	&\qX(t+1) = \max \big( 0 , \qX(t) + \auxilaryX(t) -
	\sum_{\user} \knowledgevec_\user\transpose(t) \resourceallocationvec_\user(t) \big).
	\end{eqnarray}
\end{subequations}
Therefore, $ \eqref{eq:modified_problem}$ can be recast as follows:
\begin{subequations}\label{eq:lypunov_problem}
	\begin{align}
	\label{obj:lyapunov_problem}
	\underset{\substack{[\Delta \boldsymbol{\dualvar}_{\user}(t)]_{\user}, \scheduledvec(t), \resourceallocationMat(t), \auxilary(t), \auxilaryX(t) \forall t}}{\text{minimize}} 
	& \datasetsize (1-\bar{\auxilary})^\trainingDuration 
	- \coeffKnowledge\bar{\auxilaryX}, \\
	\label{cns:lyapunov_constrains}
	\text{subject to \hphantom{00}} 
	& \eqref{cns:all_modified}, \eqref{eq:queue}, \\
	\label{cns:aux_var}
	& 0 \leq \auxilary(t) \leq 1-\beta \quad \forall t, \\
	\label{cns:auxX_var}
	& 0 \leq \auxilaryX(t) \leq \auxilaryX_0 \quad \forall t, \\
	\label{cns:utilization}
	& \textstyle u(t) = \sum_\user \frac{(1-\beta)\datasetsize_\user}{\datasetsize}\scheduled_\user(t) \quad \forall t.
	\end{align}
\end{subequations}

The quadratic Lyapunov function of $ (\q(t), \qX(t)) $ is $ L (t) = {\big(\q(t)^2 + \qX(t)^2\big)}/{2} $.
Given $\big(\q(t),\qX(t)\big)$, the expected conditional Lyapunov one slot drift at time $ t $ is $ \Delta L = \mathbb{E}[L(t+1)-L(t)|\q(t),\qX(t)] $.
Weighted by a trade-off parameter $ \trade(\geq 0) $, we add a penalty term to penalize a deviation from the optimal solution to obtain the Lyapunov DPP \cite{neely2010stochastic},
\begin{multline}
\label{pen}
\trade \big( 
\frac{\partial }{\partial \auxilary} [(1-{\auxilary})^T\datasetsize]_{\auxilary=\tilde{\auxilary}(t)} \mathbb{E}[\auxilary (t)  | \q(t)]  
- \coeffKnowledge \mathbb{E}[\auxilaryX (t)  | \qX(t)] 
\big) = \\
-\trade \Big( 
\datasetsize\trainingDuration\big(1-\tilde{\auxilary}(t)\big)^{\trainingDuration-1} \mathbb{E}[\auxilary (t)  | \q(t)]
+ \coeffKnowledge \mathbb{E}[\auxilaryX (t)  | \qX(t)] 
\Big),
\end{multline}
Here, $\tilde{\auxilary}(t) = \frac{1}{t}\sum_{\tau=1}^t \auxilary(\tau)$ and $\tilde{\auxilaryX}(t) = \frac{1}{t}\sum_{\tau=1}^t \auxilaryX(\tau)$ are the running time averages of the auxiliary variables at time $t$.
\begin{theorem}
	\label{theorem:Lyapunovupperbound}
	The upper bound of the Lyapunov DPP is given by,
	\begin{multline}\label{eq:DPP}
	\Delta L
	-\trade \Big( 
	\datasetsize\trainingDuration\big(1-\tilde{\auxilary}(t)\big)^{\trainingDuration-1} \mathbb{E}[\auxilary (t)  | \q(t)]
	+ \coeffKnowledge \mathbb{E}[\auxilaryX (t)  | \qX(t)] 
	\Big)
	\leq \\
	\textstyle  
	\mathbb{E}[\q(t)\big( \auxilary(t) - u(t) \big) 
	+ \qX(t)\big( \auxilaryX(t) - \sum_{\user} \knowledgevec_\user\transpose(t) \resourceallocationvec_\user(t) \big)
	+ L_0 \\
	-\trade \Big( 
	\datasetsize\trainingDuration\big(1-\tilde{\auxilary}(t)\big)^{\trainingDuration-1} \auxilary (t) 
	+ \coeffKnowledge \auxilaryX(t) \Big)
	| \q(t),\qX(t)],
	\end{multline}
\end{theorem}
\begin{IEEEproof}
	See Appendix \ref{Lyapunovupperbound}.
\end{IEEEproof}
The motivation behind deriving the Lyapunov \gls{dpp} is that minimizing the upper bound of the expected conditional Lyapunov \gls{dpp} at each iteration $t$ with a predefined $\trade$ yields the tradeoff between the virtual queue stability and the optimality of the solution for \eqref{eq:lypunov_problem} \cite{neely2010stochastic}.
In this regard, the stochastic optimization problem of \eqref{eq:lypunov_problem} is solved via minimizing the upper bound in \eqref{eq:DPP} at each time $t$ as follows:
\begin{subequations}\label{eq:per_tbeforerelax}
	\begin{align}
	\label{objective:first}
	\underset{\scheduledvec(t), \resourceallocationMat(t), \auxilary(t), \auxilaryX(t) }{\text{maximize}} \hphantom{0}
	& \textstyle
	\sum_\user \big(
	\frac{\q(t)(1-\beta)\datasetsize_\user }{\datasetsize}\scheduled_\user(t)
	+ \qX(t)\knowledgevec_\user\transpose(t) \resourceallocationvec_\user(t)
	\big) -  \nonumber \\ &\qquad \chi(t)\auxilary(t) 
	- \big( \qX(t) - \trade\coeffKnowledge \big) \auxilaryX(t),
	\\
	\label{cns:per_tbeforerelax}
	\text{subject to} \hphantom{00}
	& \eqref{cns:model_constraints}\text{-}\eqref{cns:RB_availability}, \eqref{cns:aux_var}, \eqref{cns:auxX_var}, \\
	\label{cns:constrains}
	& {\scheduledvec(t) \in \{0,1\}^\numberofusers,\resourceallocationvec_\user(t) \in \{0,1\}^\resourceblock \quad \forall t}, 
	\end{align}
\end{subequations}
where 
$\chi(t) = \q(t) - \trade\datasetsize\trainingDuration\big(1-\tilde{\auxilary}(t)\big)^{\trainingDuration-1}$ and the variables $\Delta \boldsymbol{\dualvar}_\user(t)$ with constraints \eqref{eq:delta_theta} and \eqref{model_update_new} are decoupled from \eqref{eq:per_tbeforerelax}. {By relaxing the integer (more specifically, boolean) variables in \eqref{cns:constrains} as linear variables, the objective and constraints become affine, in which, \eqref{eq:per_tbeforerelax} is recast as a \gls{lp} as follows:}
\begin{subequations}\label{eq:problem_per_t}
	\begin{align}
	\label{objective:problem_per_t}
	\underset{\scheduledvec(t), \resourceallocationMat(t), \auxilary(t), \auxilaryX(t) }{\text{maximize}} \hphantom{0}
	& \textstyle
	\sum_\user \big(
	\frac{\q(t)(1-\beta)\datasetsize_\user }{\datasetsize}\scheduled_\user(t)
	+ \qX(t)\knowledgevec_\user\transpose(t) \resourceallocationvec_\user(t)
	\big) -  \nonumber \\ &\qquad \chi(t)\auxilary(t) 
	- \big( \qX(t) - \trade\coeffKnowledge \big) \auxilaryX(t),
	\\
	\label{cns:per_t}
	\text{subject to} \hphantom{00}
	& \eqref{cns:model_constraints}\text{-}\eqref{cns:RB_availability}, \eqref{cns:aux_var}, \eqref{cns:auxX_var}, \\
	\label{cns:relaxed_constrains}
	& \zero \preceq \scheduledvec(t),\resourceallocationvec_\user(t) \preceq \one.
	\end{align}
\end{subequations}
Due to the independence, the optimal auxiliary variables are derived by decoupling \eqref{obj:lyapunov_problem}, \eqref{cns:aux_var}, and \eqref{cns:auxX_var} as follows:
\begin{equation}\label{eq:auxiliary_optimal}
\auxilary\optimal(t) =
\begin{cases}
1-\beta & \text{if}~ \chi(t) \geq 0, \\
0 & \text{otherwise},
\end{cases}
\quad 
\auxilaryX\optimal(t) =
\begin{cases}
\auxilaryX_0  & \text{if}~ \qX(t) \geq \trade\coeffKnowledge, \\
0 & \text{otherwise}.
\end{cases}
\end{equation}

\begin{theorem}
	\label{theorm:ipm}
	\Copy{theoremipm}{
	{
	The optimal scheduling $\scheduledvec\optimal(t)$ and \gls{rb} allocation variables $\resourceallocationMat\optimal(t)$ are found using an \gls{ipm}.}}
\end{theorem}

\begin{IEEEproof}
	{See Appendix \ref{ipm2}.}
\end{IEEEproof}

\begin{figure}[H]
	\centering
	\includegraphics[width = \columnwidth]{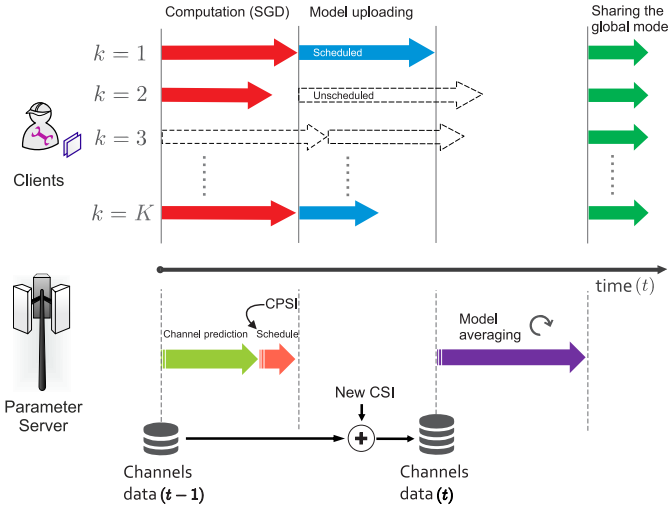}
	\caption{Illustration of the flow of the operation per client and \gls{ps} over the iterative training process.}
	\label{fig:flow_diagram}
\end{figure}

The joint client scheduling and \gls{rb} allocation is summarized in Algorithm \ref{alg:schedluing_RB_allocation} and the iterative procedure is illustrated in Fig. \ref{fig:flow_diagram}.
\begin{algorithm}[!t]
	\caption{Joint Client Scheduling and RB Allocation}
	\label{alg:schedluing_RB_allocation}
	\begin{algorithmic}[1]
		\renewcommand{\algorithmicrequire}{\textbf{Input:}}
		\renewcommand{\algorithmicensure}{\textbf{Output:}}
		\REQUIRE $ \dataset,\sinrTH, \beta, \power, \numberofresourceblocks, \xi $
		\ENSURE  $ \scheduledvec\optimal(t), \resourceallocationMat\optimal(t) $ for all $t$
		\STATE $ \q(0) =\qX(0)=0 $, $ \auxilary(0)=\auxilaryX(0) = 0 $,  $\boldsymbol{v}(0)=\zero$
		\FOR {$t = 1$ to $\trainingDuration$}
		\STATE Each client update \gls{ps} with \gls{psi}
		\STATE Each client computes $\Delta\boldsymbol{\dualvar}_\user(t)$ using \eqref{eq:delta_theta}
		\STATE Channel prediction using \gls{gpr} with \eqref{eq:gpr_mean}
		\STATE Calculate $\auxilary\optimal(t)$ and $\auxilaryX\optimal(t)$ using \eqref{eq:auxiliary_optimal}
		\STATE Derive $\scheduledvec\optimal(t)$ and $\resourceallocationMat\optimal(t)$ by solving \eqref{eq:problem_per_t} using an IPM
		\STATE Local \gls{msi} $( \Delta \boldsymbol{v}_\user(t),  \Delta \boldsymbol{\dualvar}_\user(t))$ uploading to the server
		\STATE Update $\tilde{\auxilary}(t)$, $\q(t)$ via \eqref{eq:queue}, $\boldsymbol{v}(t)$ and $\boldsymbol{\dualvar}(t)$ with \eqref{model_update_new}
		\STATE Global model $\boldsymbol{v}(t)$ broadcasting
		\STATE $t \to t+1 $
		\ENDFOR
	\end{algorithmic}
\end{algorithm}
First, all clients compute their local models using local \gls{sgd} iterations with available computation power and upload the models to the \gls{ps}. Parallelly, at the \gls{ps}, 
the channels are predicted using \gls{gpr} based on prior \gls{csi} samples and clients are scheduled following the scheduling policy shown in Algorithm \ref{alg:schedluing_RB_allocation}. Scheduled clients upload their local models to the \gls{ps}, then at the \gls{ps} the received models are averaged out to obtain a new global model and broadcast back to all $ \numberofusers $ clients. In this setting, by sampling the scheduled clients, the \gls{ps} gets additional information on the \gls{csi}.
In the example presented in Fig. \ref{fig:flow_diagram}, the dashed arrows correspond to non-scheduled clients due to the lack of computational resources and/or poor channel conditions.
This strategy allows the proposed scheduling method to avoid unnecessary computation and communication delays.

{\subsection{Convergence, optimality, and complexity of the proposed solution}
	\Copy{Wholetext}{\Copy{proofConvergence}{Towards solving \eqref{obj:master_problem}, the main objective \eqref{obj:master_problem} is decoupled over clients and server using a dual formulation.
	Then, the upper bound of $ \e(T) $, which is the accuracy loss using scheduling compared to a centralized training, is derived in Theorem \ref{theorem:gap}, in which, solving \eqref{eq:modified_problem} is equivalent to solving \eqref{eq:master_problem}.	
	It is worth noting that minimizing $ \e(T) $, guarantees convergence to the minimum training loss, which is achieved as $ T \to \infty $ \cite[Appendix B]{yang2019scheduling}.	
	In \eqref{eq:modified_problem}, the minimization of the analytical expression of $ \e(T) $ with a finite $ T $ boils down to a stochastic optimization problem with a nonlinear time average objective.
	The stochastic optimization problem \eqref{eq:modified_problem} is decoupled into a sequence of optimization problems \eqref{eq:per_tbeforerelax} that are solved at each time iteration $ t $ using the Lyapunov \gls{dpp} technique with guaranteed convergence \cite[Section 4]{neely2012stability}.
	Following Remark 1, the solution of \eqref{eq:problem_per_t} yields the optimality of \eqref{eq:per_tbeforerelax} ensuring that the convergence guarantees are held under the Lypunov \gls{dpp} method.} \Copy{proofOptimality}{Since the optimal solution of \eqref{eq:problem_per_t} is optimal for \eqref{eq:per_tbeforerelax}, the optimality of the proposed solution depends on the recasted problem \eqref{eq:lypunov_problem} that relies on the Lypunov \gls{dpp} technique.
	Moreover, the optimality of the Lyapunov \gls{dpp}-based solution is in the order of $ \mathcal{O}(\frac{1}{\coeffKnowledge}) $ \cite[Section 7.4]{neely2012stability}.}	
	
	Finally, \Copy{proofComplexity}{the complexity of Algorithm 1 depends on the complexity of the \gls{ipm} used to solve the \gls{lp}. Specifically, the complexity of the proposed solution at each iteration is given by the computational complexity of solving the \gls{lp} which is in the order of $ \mathcal{O}(n^{3}L) $ \cite{gritzmann1994complexity} with $ n =(\datasetsize + \numberofresourceblocks + 1)\numberofusers +2 $ variables, each of which is represented by a $ L $-bits code.}}
}

\section{Simulation Results}
\label{res}

\begin{table}[t]
    \caption{Simulation parameters }
    \label{tableofparameters}
    \begin{center}
        \begin{tabular}{l c}
        \hline
          \multicolumn{1}{c}{\textbf{Parameter}}
        & \multicolumn{1}{c}{\textbf{Value}}
        \\
        \hline
        \hline
		  Number of clients				 ($\numberofusers$)			&$10$   \\
          Number of \gls{rb}			 ($\numberofresourceblocks$)&$6$    \\
          Local model solving optimality ($\beta$)        			&$0.7$  \\
          Transmit Power (in Watts)      ($\power$) 			&$ 1 $ \\          
          \hline
          \multicolumn{2}{c}{\textbf{Model training and scheduling}}\\
          \hline
          Training duration				 ($\trainingDuration$) 		&$ 100 $\\
          Number of local \gls{sgd} iterations ($\numberoflocaliterations$)&$10$  \\
          Learning rate					 ($\eta  $)       			&$0.2$  \\
          Regularizer parameter			 ($ \xi  $)   				&$ 1$   \\  
          Lyapunov trade-off parameter	 ($\trade$)       			&$1$    \\ 
          Tradeoff weight	             ($\coeffKnowledge$)  		&$ 1 $  \\
          SINR threshold 				 ($\sinrTH$)      			&$1.2$  \\
          Computation time threshold	 ($ \computationtime_0 $) 	&$ 1.2 $\\
          \hline
          \multicolumn{2}{c}{\textbf{Channel prediction (\gls{gpr})}}\\
          \hline
          length parameter	 ($\gprLength$)   			&$ 2 $ 	\\
          period parameter	 ($\gprPeriod$)   			&$5$    \\
          Number of past observations ($ N $)  			&$20$   \\
          \hline
        \end{tabular}
    \end{center}
\end{table}

In this section, we evaluate the proposed client scheduling method and \gls{rb} allocation using MNIST and CIFAR-10 datasets assuming $\lossfunc(\cdot)$ and $\re(\cdot)$ as the cross entropy loss function and Tikhonov regularizer, respectively.
A subset of the MNIST dataset with 6000 samples consisting of equal sizes of ten classes of 0-9 digits are distributed over $\numberofusers=10$ clients, whereas for CIFAR-10 data samples are from ten different categories but following the same data distribution. In addition, the wireless channel follows a correlated Rayleigh distribution \cite{Rappaport1996} with mean to noise ratio equal to $\sinrTH$.
For perfect \gls{csi}, a single \gls{rb} is dedicated for channel estimation.
The remaining parameters are presented in Table \ref{tableofparameters}.
\begin{figure}[!t]
	\centering
	\includegraphics[width=\linewidth]{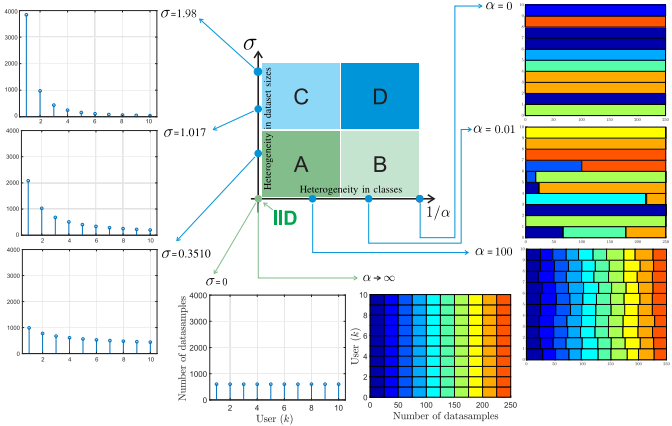}
	\caption{From \gls{IID} datasets to non-\gls{IID} datasets under different choices of Zipf parameter $ (\sigma) $ and Dirichlet parameter $( \diricheletparameter)$. The performance of the proposed algorithms are evaluated under the choices highlighted with the four regions A, B, C, and D.}
	\label{iiddistribution}
\end{figure}

For \gls{IID} datasets, training data is partitioned into $\numberofusers$ subsets of equal sizes with each consisting of equal number of samples from all $ 10 $ classes, which are randomly distributed over the $\numberofusers$ clients.
The impact of the non-\gls{IID} on the performance is studied for i) \emph{heterogeneous dataset sizes}: clients having training datasets with different number of samples  and ii) \emph{heterogeneous class availability}: clients' datasets contain different number of samples per class.
The dataset size heterogeneity is modeled by partitioning the training dataset over clients using the Zipf distribution, in which, the dataset of client $\user$ is composed of $\datasetsize_\user = \datasetsize\user^{-\sigma}/\sum_{\varkappa\in\userSet} \varkappa^{-\sigma}$ number of samples \cite{moreno2016large}.
Here, the Zipf's parameter $\sigma=0$ yields uniform/homogeneous data distribution over clients (600 samples per client), whereas increasing $\sigma$ results in heterogeneous dataset sizes among clients as shown in Fig. \ref{iiddistribution}.
To control the heterogeneity in class availability over clients, we adopt the Dirichelet distribution, which is parameterized by a concentration parameter $ \diricheletparameter \in (0,\infty] $ to distribute data samples from each class among clients \cite{pitman1997two}.
With $ \diricheletparameter = 0 $, each client's dataset consists of samples from a single class in which increasing alpha yields datasets with training data from several classes but the majority of data is from few classes. As $ \diricheletparameter \to \infty$, each client receives a dataset with samples drawn uniformly from all classes as illustrated in Fig. \ref{iiddistribution}.

Throughout the discussion, \emph{centralized training} refers to training that takes place at the \gls{ps} with access to the entire dataset.
In addition, how well the global model generalizes for individual clients is termed as ``generalization" in this discussion.
Training data samples are distributed among clients except in centralized training.
In the proposed approach data samples are drawn from ten classes of handwritten digits. We compare several proposed \gls{rb} allocation and client scheduling policies as well as other baseline methods. Under perfect \gls{csi}, two variants of the proposed methods named \emph{quantity-aware} scheduling \propPerfect{} and \emph{quantity-unaware} scheduling \basePaper{} are compared, whose difference stems from either accounting or neglecting the local dataset size in model updates during scheduling. Under imperfect \gls{csi}, \gls{gpr}-based channel prediction is combined together with \propPerfect{} yielding the \propImperfect{} method. For comparison, we adopt two baselines: a random scheduling technique \baseRand{} and a proportional fair \basePF{} method where fairness is expressed in terms of successful model uploading.
In addition, we use the vanilla FL method \cite{konevcny2016federated} without \gls{rb} constraints, denoted as \vanilaFL{} hereinafter. {All proposed scheduling policies as well as baseline methods are summarized in Table \ref{tab:simulations}.
\begin{table}[ht]
	\centering
	\renewcommand\multirowsetup{\centering}
	\caption{Proposed Algorithms and benchmark algorithm}
	\label{tab:simulations}
	\begin{tabular}{p{1.5 cm} p{1.2 cm} p{4 cm}}
			\cmidrule[0.75 pt]{2-3}
			& \makebox[1.2 cm][c]{{\bf Model}} & \makebox[4 cm][c]{{\bf Description}} \\
			\cmidrule[0.75 pt]{2-3}			
			\ldelim \{ {9}{1.2 cm}[\parbox{1.2 cm}{Proposed Methods}]
			&
			\propImperfect{} &  PS uses dataset sizes to prioritize clients and \gls{gpr}-based \gls{csi} predictions for scheduling.\\
			& 
			\propPerfect{} & \gls{ps} uses pilot-based \gls{csi} estimation and dataset sizes to prioritize clients. \\
			&
			\basePaper{} & \gls{ps} uses pilot-based \gls{csi} estimation without accounting dataset size of clients. \\
			\cline{2-3}			
			\ldelim \{ {4.2}{1.2 cm}[\parbox{1.2 cm}{Baselines}]
			& \vspace{0em} \baseRand{} & \vspace{0em} Clients are scheduled randomly.\\
			&\basePF{} & Clients are scheduled with fairness in terms of successful model uploading.\\
			\cline{2-3}		
			\ldelim \{ {3.8}{1.2 cm}[\parbox{1.2 cm}{Ideal setup}]			
			& \vspace{0em}\vanilaFL{} & \vspace{0em} All clients are scheduled assuming no communication or computation constraints.\\
			\cmidrule[0.75 pt]{2-3}
	\end{tabular}
\end{table}
}
\subsection{Loss of accuracy}
\begin{figure}
	\begin{subfigure}{\columnwidth}
		\centering\includegraphics[width=\linewidth]{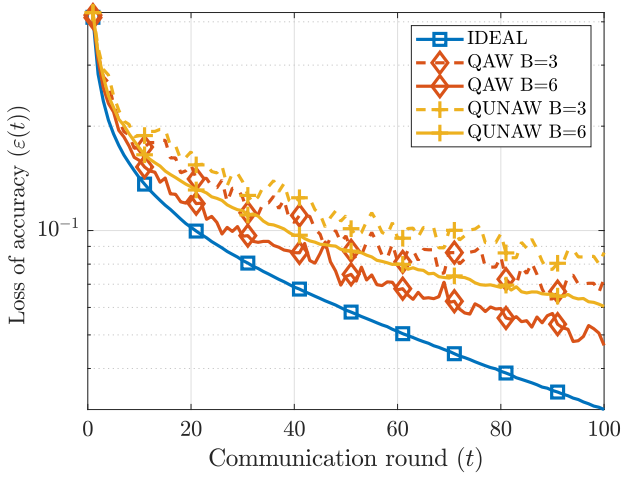}
		\caption{FL with perfect \gls{csi} and $\numberofresourceblocks=\{3,6\}$.}
		\label{fig:epsilon_perfect}
	\end{subfigure}
	\begin{subfigure}{\columnwidth}
		\centering\includegraphics[width=\linewidth]{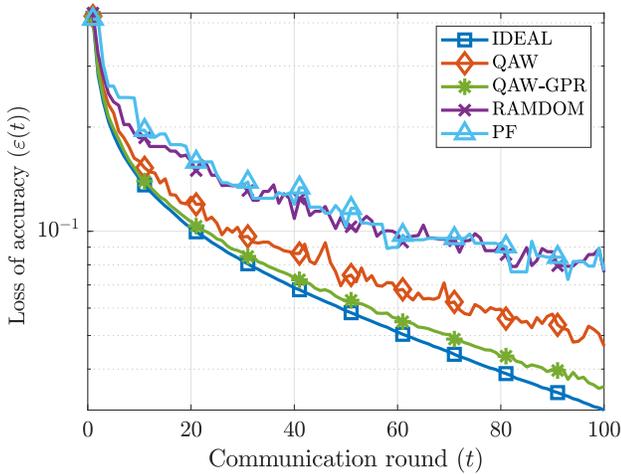}
		\caption{FL with imperfect \gls{csi} and $\numberofresourceblocks=6$.}
		\label{fig:epsilon_imperfect}
	\end{subfigure}
	\caption{Comparison of the loss of accuracy in all FL methods for each model aggregation round vs. centralized training, Zipf parameter $\sigma=1.017$ and $ \diricheletparameter \to \infty $(region C of Fig. \ref{iiddistribution}).}
	\label{fig:epsillon}
\end{figure}

Fig. \ref{fig:epsillon} compares the loss of accuracy in all \gls{fl} methods at each model aggregation round with respect to the centralized model training.
Here, we have considered the unbalanced dataset distribution ($\sigma=1.017$) among clients to analyze the impact of dataset size in scheduling.
It can be noted that \vanilaFL{} has the lowest loss of accuracy $\e(100)=0.03$ due to the absence of both communication and computation constraints.
Under perfect \gls{csi}, Fig. \ref{fig:epsilon_perfect} plots \propPerfect{} and \basePaper{} for two different \gls{rb} values $\numberofresourceblocks \in \lbrace 3,6\rbrace $.
With a 2$\times$ increase in \gls{rb}s, the gain of the gap in loss in both \propPerfect{} and \basePaper{} is almost the same.
For $\numberofresourceblocks =6$, Fig. \ref{fig:epsilon_perfect} shows that the \propPerfect{} reduces the gap in loss by $ 22.8\,\%$ compared to \basePaper{}. The reason for that is that \propPerfect{} cleverly schedules clients with higher data samples compared to \basePaper{} when the dataset distribution among clients is unbalanced.
Under imperfect \gls{csi}, \propImperfect{}, \basePF{}, and \baseRand{} are compared in Fig. \ref{fig:epsilon_imperfect} alongside $\vanilaFL{}$ and $\propPerfect{}$.
While \baseRand{} and \basePF{} show a poor performance, \propImperfect{} outperforms \propPerfect{} by reducing the gap in loss by $23.6\,\%$.
The main reason for this improvement is that \propPerfect{} needs to sacrifice some of its \gls{rb}s for channel measurements while \gls{csi} prediction in \propImperfect{} leverages all \gls{rb}s.

\subsection{Impact of system parameters}
\begin{figure}[t]
	\centering
	\includegraphics[width= \columnwidth]{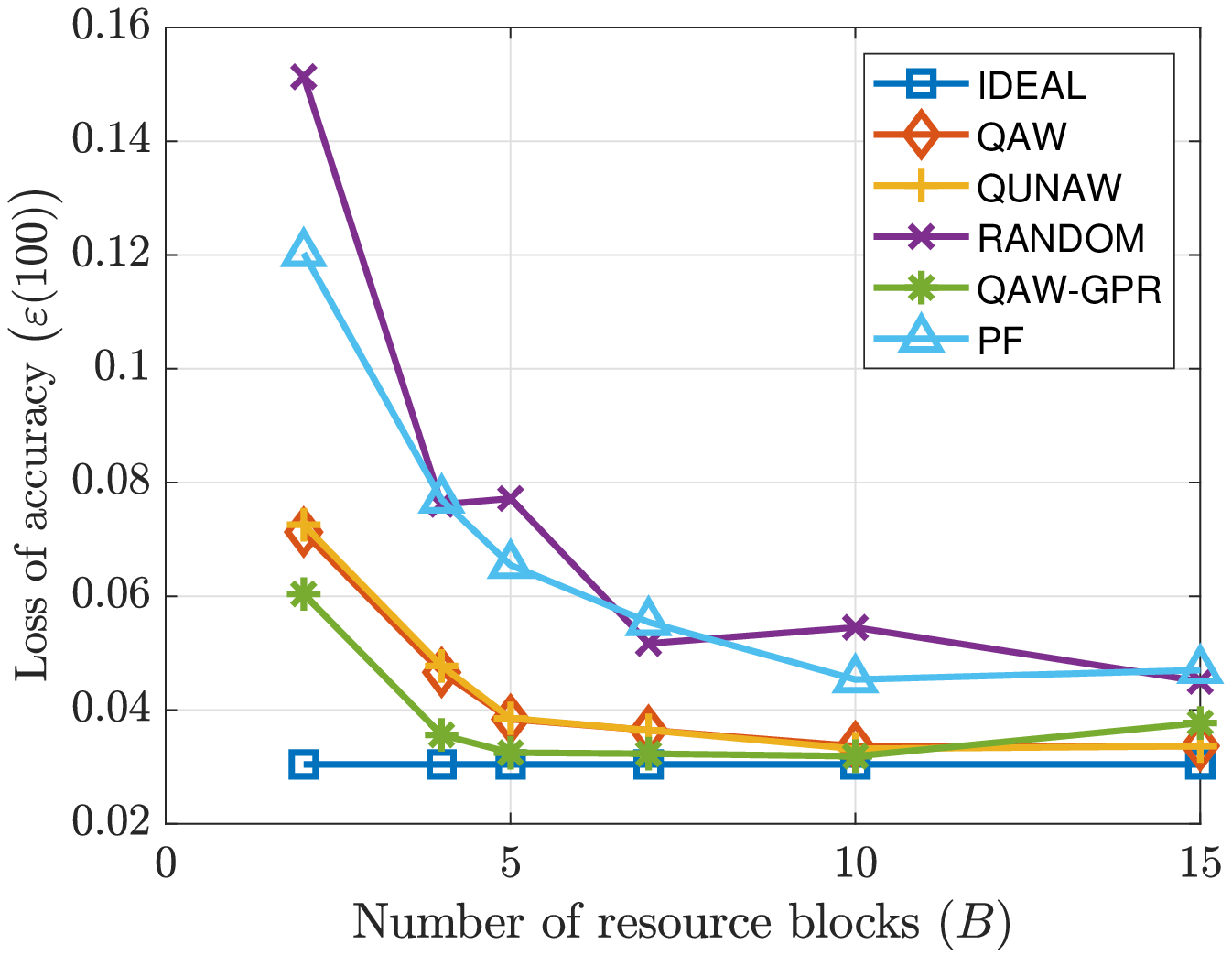}
	\caption{Impact of the available \gls{rb}s on the gap of loss $ \e(100) $ for $ \numberofusers = 10 $ clients for $ \sigma=0 $ and $ \diricheletparameter \to \infty $ dataset distribution (region A of Fig. \ref{iiddistribution}).}
	\label{fig:resourceblock}
\end{figure}

Fig. \ref{fig:resourceblock} shows the impact of the available communication resources (\gls{rb}s) on the performance of the trained models $ F(\modelvec(100),\mathcal{\dataset}) $.
Without computing and communication constraints, \vanilaFL{} shows the lowest loss while \baseRand{} and \basePF{} exhibit the highest losses due to client scheduling with limited \gls{rb}s.
On the other hand the proposed methods \propImperfect{}, performs better than \propPerfect{} and \basePaper{} for $ \numberofresourceblocks \leq 10$ thanks to additional \gls{rb}s when \gls{csi} measurements are missing.
Beyond $ \numberofresourceblocks = 10 $, the available number of \gls{rb}s exceeds $ \numberofusers $, and thus, channel sampling in \propImperfect{} is limited to at most $ \numberofusers $ samples.
Hence, increasing $ \numberofresourceblocks $ beyond $ \numberofusers = 10 $ results in increased number of under-sampled \gls{rb}s yielding high uncertainty in \gls{gpr} and poor \gls{csi} predictions. Inaccurate \gls{csi} prediction leads to scheduling clients with weak channels (stragglers), which consequently provides a loss of performance in \propImperfect{}.

\begin{figure}[!t]
	\centering
	\includegraphics[width=\columnwidth]{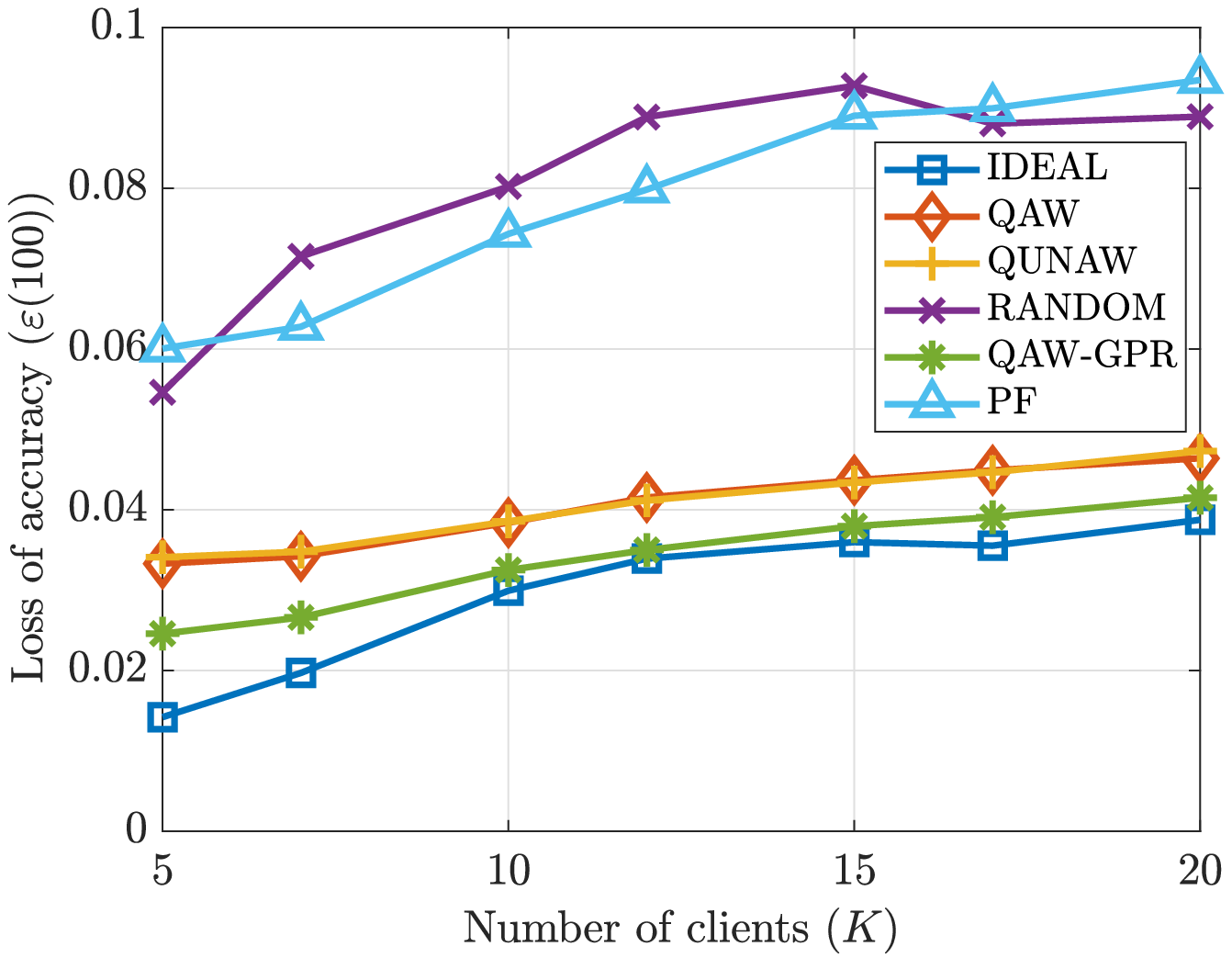}
	\caption{Impact of the number of clients on the gap of accuracy loss $ \e(100) $ for $ \numberofresourceblocks = 5 $ \gls{rb}s for $ \sigma=0 $ and $ \diricheletparameter \to \infty $ dataset distribution (region A of Fig. \ref{iiddistribution}).}
	\label{fig:userplot}
\end{figure}

The impact of the number of clients ($ \numberofusers $) in the system with fixed \gls{rb}s ($ \numberofresourceblocks = 5 $) on the trained models performance and under different training policies is shown in Fig. \ref{fig:userplot}.
It can be noted that all methods exhibit higher losses when increasing $ \numberofusers $ due to: i) local training with fewer data samples ($ 2500/\numberofusers $) which deteriorates in the non-\gls{IID} regime, ii) the limited fraction of clients ($ 5/ \numberofusers $) that are scheduled at once (except for the \vanilaFL{} method).
The choice of equal number of samples per clients (balanced data sets with $ \sigma = 0 $) results highlights that \propPerfect{} and \basePaper{} exhibit identical performance as shown in Fig. \ref{fig:userplot}.
Under limited resources, \propImperfect{} outperforms all other proposed methods and baselines by reaping the benefits of additional \gls{rb}s for \gls{gpr}-based channel prediction.
The baseline methods \basePF{} and \baseRand{} are oblivious to both \gls{csi} and training performance, giving rise to higher losses compared to the proposed methods.

\begin{figure}[!t]
	\centering
	\includegraphics[width= \columnwidth]{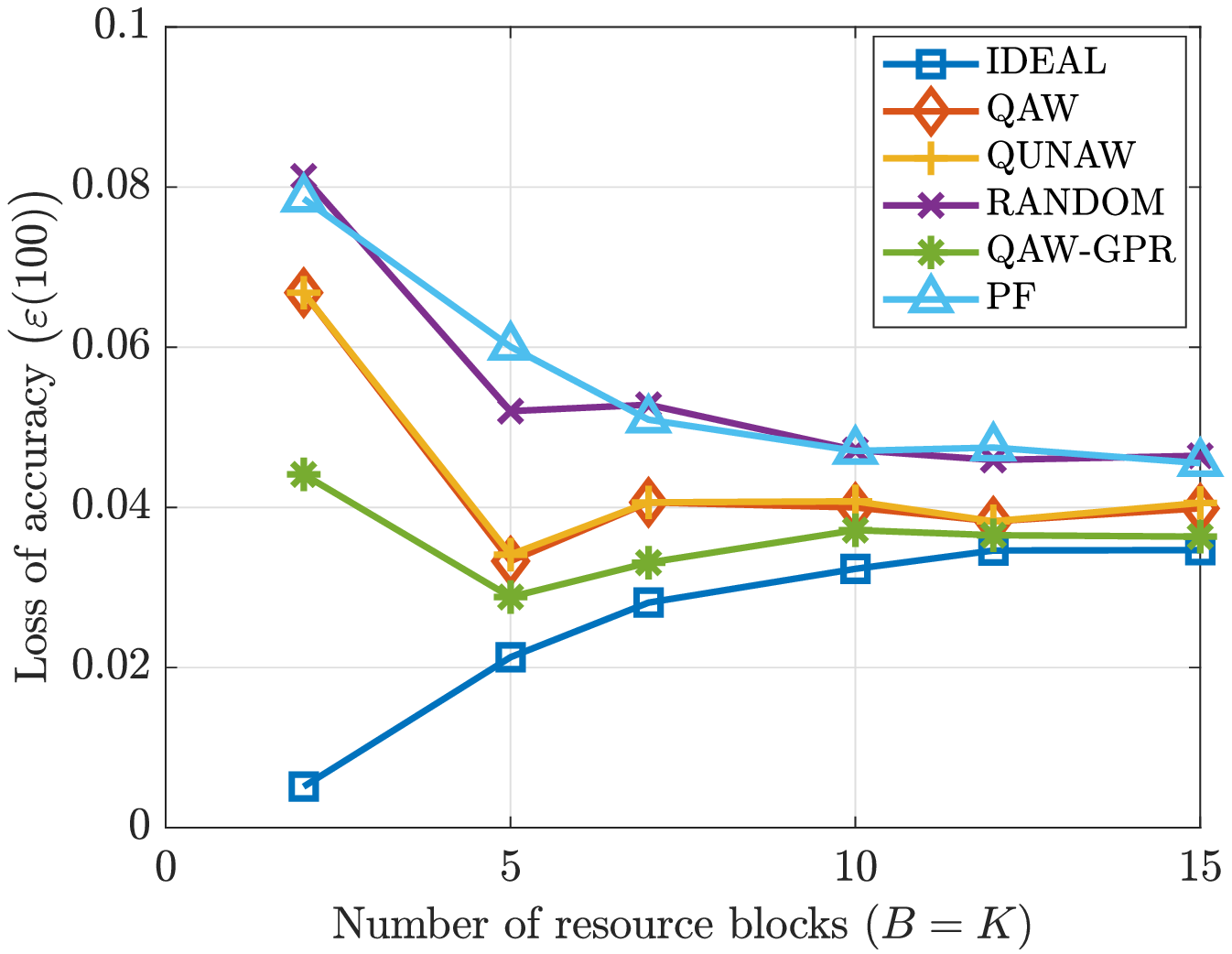}
	\caption{The analysis of the loss of accuracy $ \e(100) $ as the system scales with fixed clients to \gls{rb}s ratio, (i.e., $ \numberofresourceblocks = \numberofusers$), with $ \sigma=0 $ and $ \diricheletparameter \to \infty $ dataset distribution (region A of Fig. \ref{iiddistribution}).}
	\label{fig:rbuser}
\end{figure}

Fig. \ref{fig:rbuser} analyzes the performance as the network scales uniformly with the number of \gls{rb}s and users, i.e., $ \numberofusers = \numberofresourceblocks $.
Under ideal conditions, the loss in performance when increasing $ \numberofusers $ shown in Fig. \ref{fig:rbuser} follows the same reasoning presented under Fig. \ref{fig:userplot}.
The \gls{gpr}-based \gls{csi} prediction allows \propImperfect{} to allocate all \gls{rb}s for client scheduling yielding the best performance out of the proposed and baseline methods.
It is also shown that the worst performance among all methods except \vanilaFL{} is at $ \numberofusers = \numberofresourceblocks = 2 $, owing to the dynamics of \gls{csi} leading to scheduling stragglers over limited \gls{rb}s.
With increasing $ \numberofusers $ and $ \numberofresourceblocks $, the number of possibilities that clients can be successfully scheduled increases, leading to improved performance in all baseline and proposed methods. In contrast to the baseline methods, the proposed methods optimize client scheduling to reduce the loss of accuracy, achieving closer performance to \vanilaFL{} with increasing K and B. However, beyond $ \numberofusers = \numberofresourceblocks = 5 $, the performance loss due to smaller local datasets outweighs the performance gains coming from an increasing number of scheduled clients, and thus, all three proposed methods follow the trend of \vanilaFL{} as illustrated in Fig. \ref{fig:rbuser}.
Compared to the \basePF{} baseline with $ \numberofusers = \numberofresourceblocks = 15 $, \propImperfect{} shows a reduction in the loss of accuracy by $ 20\,\% $.

\subsection{Impact of dataset distribution}
\begin{figure}[!t]
	\centering
	\includegraphics[width=\columnwidth]{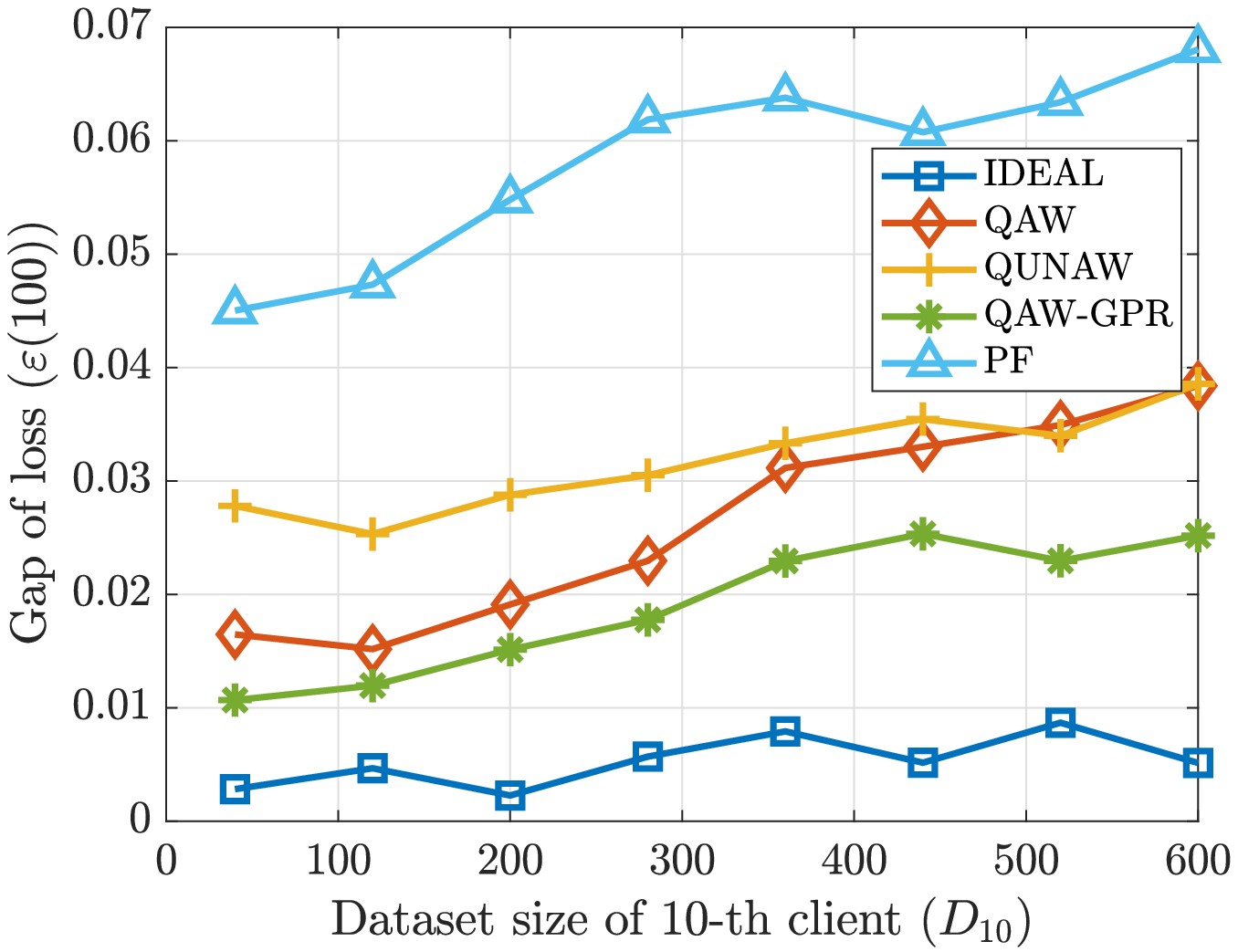}
	\caption{Impact of dataset distribution balanced-unbalancedness on the loss of accuracy $ \e(100) $ for $ \numberofresourceblocks = 5 $, $ \numberofusers = 10 $, with $ \diricheletparameter \to \infty $.}
	\label{fig:accuracy}
\end{figure}

Fig. \ref{fig:accuracy} plots the impact of data distribution in terms of balanced-unbalancedness in terms of training sample size per client on the loss of accuracy $ \e(100) $.
Here, the $x$-axis represents the local dataset size of the client having the lowest number of training data, i.e., the dataset size of the 10th client $\datasetsize_{10}$ as per the Zipf distribution with $ \diricheletparameter \to \infty $.
With balanced datasets, all clients equally contribute to model training, hence scheduling a fraction of the clients results in a significant loss in performance. In contrast, differences in dataset sizes reflect the importance of clients with large datasets.
Therefore, scheduling important clients yields lower gaps in performance, even for methods that are oblivious to dataset sizes but fairly schedule all clients, as shown in Fig. \ref{fig:accuracy}.
Among the proposed methods, \propImperfect{} outperforms the others thanks to using additional \gls{rb}s with the absence of \gls{csi} measurement.
Compared to the baselines \basePF{} and \propPerfect{}, \propImperfect{} shows a reduction of loss of accuracy by $ 76.3\,\% $,  for the highest skewed data distribution ($\datasetsize_{10}=40$).
In contrast, \basePaper{} yields higher losses compared to \propPerfect{}, \propImperfect{} when training data is unevenly distributed among clients.
As an example, the reduction of the loss of accuracy in \propPerfect{} at $\datasetsize_{10}=40$ is $25.72\,\%$ compared to $ 40.7\,\%$ for \basePaper{}.
The reason behind these lower losses is that client scheduling takes into account the training dataset size.
For $\datasetsize_{10}=600$, due to the equal dataset sizes per client, the accuracy loss provided by \propPerfect{} and \basePaper{} are identical.
Therein, both \propPerfect{} and \basePaper{} exhibit about $43.6\,\%$ reduction in accuracy loss compared to \baseRand{}.

\begin{figure}[t]
	\centering
	\includegraphics[width=\columnwidth]{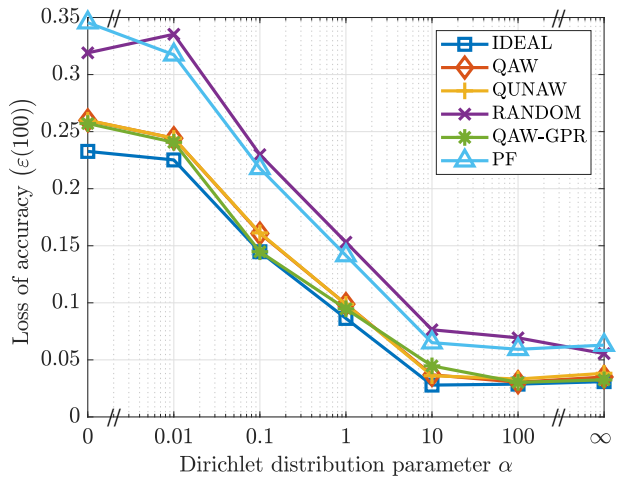}
	\caption{Impact of the class-wise data heterogeneity on the loss of accuracy with $ \sigma = 0 $.}
	\label{alpha}
\end{figure}

Next in Fig. \ref{alpha}, we analyze the impact of non-\gls{IID} data in terms of the available number of training data from all classes on the accuracy of the trained model.
Here, we compare the performance of the proposed methods with the baselines for several choices of $ \diricheletparameter $ with $ \datasetsize_\user = 250 $ for all $ \user \in \userSet $ (i.e., $\sigma = 0$).
Fig. \ref{iiddistribution} shows the class distribution over clients for different choices of $ \diricheletparameter $ with equal dataset size $ \datasetsize_\user = 250 $.
Fig. \ref{alpha} illustrates that the training performance degrades as the samples data distribution becomes class wise heterogeneous.
For instance from the \gls{IID} case ($ \diricheletparameter = 0 $) to the case with $ \diricheletparameter \to \infty $ case, \propPerfect{} achieves a loss of accuracy reduction by $ 86.5\,\%$ compared to $ \diricheletparameter = 0 $.
However, \propPerfect{}, \propImperfect{} and \basePaper{} perform better than \baseRand{} and \basePF{}.
Finally, it can be noticed that from $ \diricheletparameter = 0.01 $ to $ \diricheletparameter = 10 $ the gap in terms of accuracy loss performance of the trained model increases rapidly for all methods.
Beyond those points, changes are small comparably to the inner points.

\subsection{Impact of computing resources}
\begin{figure}[!t]
	\centering
	\includegraphics[width=\columnwidth]{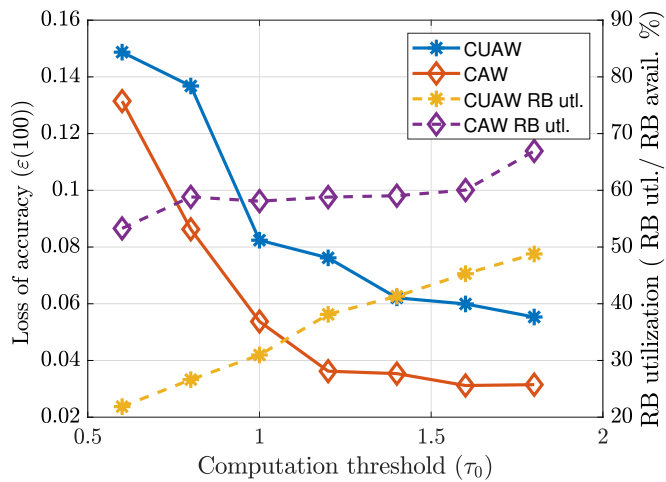}
	\caption{Impact of \gls{psi} for \propPerfect{} scheduling}
	\label{fig:computing}
\end{figure}

The impact of the limitations in computation resources in model training and client scheduling is analyzed in Fig. \ref{fig:computing}. Since allocating \gls{rb}s to stragglers results in poor \gls{rb} utilization, we define the \gls{rb} utilization metric as the percentage of \gls{rb}s used for a successful model upload over the allocated \gls{rb}s. Then we compare two variants of \propPerfect{} with and without considering the computation constraint \eqref{computationaltime} referred to as CAW (original \propPerfect{} in the previous discussions) and CUAW, respectively.
It is worth highlighting that the computation threshold is inversely proportional to the average computing power availability as per \eqref{computationtimeandclockfrequency}, i.e. lower $ \computationtime_{0} $ corresponds to higher $ \e(t) $ and vice versa. Fig. \ref{fig:computing} indicates that the use of \gls{psi} for client scheduling in CAW reduces the number of scheduled computation stragglers resulting in a lower accuracy loss, in addition to higher \gls{rb} utilization over CUAW. Overall, CAW with \propPerfect{} scheduling performs better than CUAW with \propPerfect{} scheduling. For instance, compared to CUAW, CAW achieves $11.6\,\%$ reduction in loss of accuracy with $ \computationtime_{0} =0.6 $ which increases to $43.1\,\%$ with $ \computationtime_{0} =1.4 $.
When $ \computationtime_{0} $ is small, the number of stragglers increases leading to poor performance for both CAW and CUAW.
It is worth nothing that considering \gls{psi} in the scheduling, CAW achieves at least $ 18.2\,\% $ increase in \gls{rb} utilization in addition to the loss of accuracy reduction by at least $ 11.6\,\%$.

\begin{figure}[!t]
	\centering
	\includegraphics[width=\columnwidth]{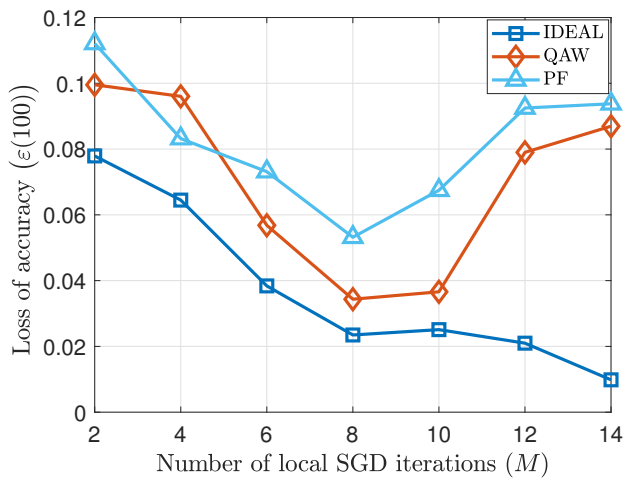}
	\caption{Impact of number of local \gls{sgd} iterations for loss of accuracy $ \e(100) $}
	\label{fig:sgditer}
\end{figure}

The impact of local \gls{sgd} iterations under limited computing power availability is analyzed in Fig. \ref{fig:sgditer}.
Due to the assumption of unlimited computing power availability, \vanilaFL{} performs well with the increasing local \gls{sgd} iterations. Similarly, gradual reductions in the loss of accuracy for both \propPerfect{} and \basePF{} can be seen as $ \numberoflocaliterations $ increases from two to eight. However, further increasing $ \numberoflocaliterations $ results in longer delays for some clients in local computing under limited processing power as per \eqref{computationtimeandclockfrequency}. Such computation stragglers do not contribute to the training in both PF (drops out due to the computation constraint) and \propPerfect{} (not scheduled). Hence, increasing local \gls{sgd} iterations beyond $ \numberoflocaliterations = 8 $ results in fewer clients to contribute for the training, in which, increased losses of accuracy with \propPerfect{} and \basePF{} are observed as shown in Fig. \ref{fig:sgditer}.

\subsection{Fairness}

\begin{figure}[!t]
	\centering
	\includegraphics[width=\columnwidth]{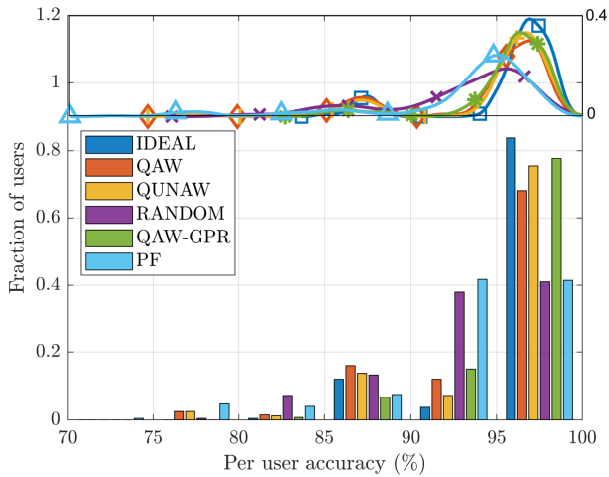}
	\caption{Fairness comparison of the training accuracy among clients, Zipf parameter $\sigma=1.071$  and $ \diricheletparameter \to \infty $ (region C of Fig. \ref{iiddistribution}).}
	\label{fig:peruseraccuracy}
\end{figure}

Fig.\ref{fig:peruseraccuracy} indicates how well the global model generalizes for individual clients.
Therein, the global model is used at the client side for inference, and the per client histogram of model accuracy is presented.
It can be seen that, the global model in \vanilaFL{} generalizes well over the clients yielding the highest accuracy on average $( 96.8\,\% )$ over all clients and the lowest variance with $ 5.6 $.
With \propImperfect{}, $ 96.1\,\% $ average accuracy and variance of $ 7.8 $ is observed.
It is also seen that \propPerfect{} and \basePaper{} have almost equal means ($ 95.2\,\% $) and variances of $ 15.2 $ and $ 16.3 $, respectively.
Scheduling clients with a larger dataset in \propPerfect{} provides a lower variance in accuracy compared to \basePaper{}.
Although \baseRand{} and \basePF{} are \gls{csi}-agnostic, they yield an average accuracy of $ 93.4\,\% $ and $ 94.1\,\% $ respectively with the highest variance of $ 18.2 $ and  $ 17.6 $.
This indicates that client scheduling without any insight on datasize distribution and \gls{csi} fails to provide high training accuracy or fairness under communication constraints.

\begin{figure}[!t]
	\begin{subfigure}{\columnwidth}
		\centering\includegraphics[width=0.95\linewidth]{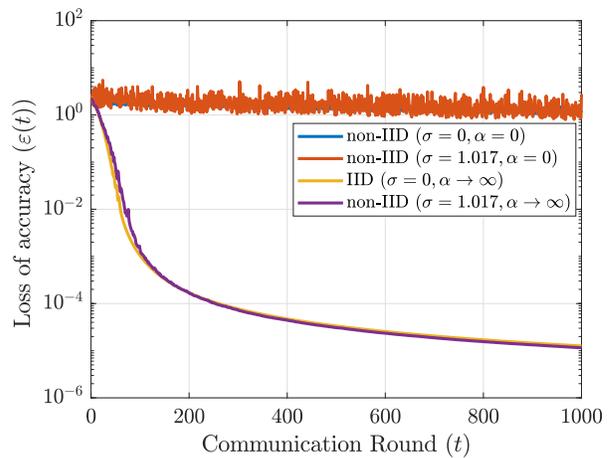}
		\caption{\vanilaFL{} scheduling}
		\label{fig:IDEAL}
	\end{subfigure}
	\begin{subfigure}{\columnwidth}
		\centering\includegraphics[width=0.95\linewidth]{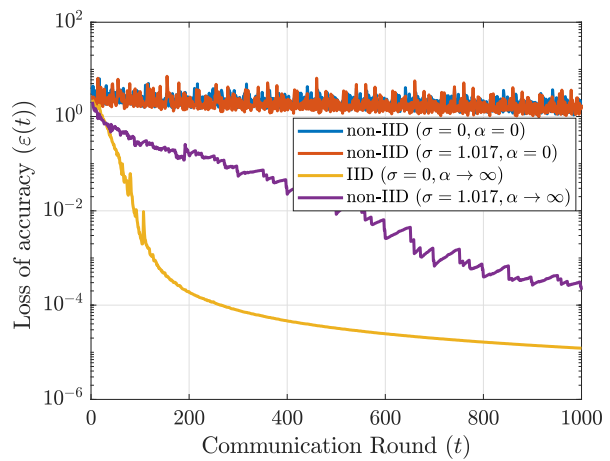}
		\caption{\propPerfect{} scheduling}
		\label{QAW}
	\end{subfigure}	
	\caption{Performance comparison of the proposed \propPerfect{} scheduling approach on CIFAR-10 trained with \gls{cnn} vs. the \vanilaFL{} scheduling scheme.}
	\label{cifar10}
\end{figure}

Finally, Fig. \ref{cifar10} compares the generalization performance of one of the proposed methods, \propPerfect{}, for CIFAR-10 dataset instead of MNIST under both \gls{IID} and non-\gls{IID} data.
In contrast to MNIST, CIFAR-10 consists of color images of physical objects from ten distinct classes \cite{krizhevsky2009learning}.
For training, we adopt a 3-layer \gls{cnn} with $ \numberofusers = 10 $ clients training over $ T = 1000 $ iterations.
A total of $ 2500 $ data samples are distributed over the clients under four settings corresponding to the four regions in Fig. \ref{iiddistribution}: i)
\gls{IID} data with $\diricheletparameter \to \infty$ and $ \sigma = 0 $ in region A, ii) equal dataset sizes $(\diricheletparameter=0)$ under heterogeneous class availability $(\diricheletparameter=0) $ in region B, iii) homogeneous class availability $(\diricheletparameter \to \infty)$ with heterogeneous dataset sizes $(\sigma = 1.017)$ in region C, and iv) heterogeneity in both $(\sigma = 1.017)$ and $(\diricheletparameter=0) $ under region D.
Fig. \ref{fig:IDEAL} indicates that the training performance with CIFAR-10 dataset significantly suffers from non-\gls{IID} data under \vanilaFL{} communication and computation conditions.
Interestingly, the proposed \propPerfect{} yields identical performance than \vanilaFL{} under balanced datasets disregarding the identicalness of the data. However, with unbalanced datasets, the training performance of \propPerfect{} is degraded as illustrated in Fig. \ref{QAW}. The underlying reason is that the unbalanced datasets induce non-\gls{IID} data at each client, and scheduling fewer clients is insufficient to obtain higher training performance.

\section{Conclusion}
\label{conclu}
In this work, \gls{fl} over wireless networks with limited computational and communication resources, and under imperfect \gls{csi} is investigated.
To achieve a training performance close to a centralized training setting, a novel client scheduling and \gls{rb} allocation policy leveraging \gls{gpr}-based channel prediction is proposed.
Through extensive sets of simulations the benefits of \gls{fl} using the proposed client scheduling and \gls{rb} allocation policy are validated and analyzed in terms of (i) system parameters, model performance and computation resource limitations (number of \gls{rb}s, number of clients) and (ii) heterogeneity of data distribution over clients (balanced-unbalanced, \gls{IID} and non-\gls{IID}).
Results show that the proposed methods reduce the gap of the accuracy by up to $ 40.7\,\%$ compared to state-of-the-art client scheduling and \gls{rb} allocation methods.

\appendix
\subsection{Proof of Theorem 1}
\label{gap_proof}
After $ t $ and $ t+1 $ communication rounds, the expected increment in the dual function of \eqref{obj:master_problem} is,
\begin{align}
	\mathbb{E}[\Dual(\boldsymbol{\dualvar}(t + 1)) - \Dual(\boldsymbol{\dualvar}(t))] & \geq \mathbb{E} [ \Dual(\boldsymbol{\dualvar}(t+1))] - 	\mathbb{E}[  \Dual(\boldsymbol{\dualvar}(t))]. 	\nonumber
\end{align}
This inequality holds since the expectations of difference is greater than the difference of the expectations of a convex function. 
By adding and subtracting the optimal dual function value to the R.H.S. :
\begin{multline}
\mathbb{E}  [\Dual(\boldsymbol{\dualvar}(t + 1)) - \Dual(\boldsymbol{\dualvar}(t))] \geq \mathbb{E} [\Dual(\boldsymbol{\dualvar}^{\star})] - \mathbb{E} [ \Dual(\boldsymbol{\dualvar}(t)) ] \nonumber \\ \qquad \qquad \qquad \qquad \qquad \qquad \qquad + 	\mathbb{E} [ \Dual(\boldsymbol{\dualvar}(t+1)) ]  - \mathbb{E} [ \Dual(\boldsymbol{\dualvar}^{\star}) ]	\nonumber \\
= \textstyle \sum_{\user = 1}^{\numberofusers}	\Delta \Dual (\Delta \boldsymbol{\dualvar}_{\user}^{\star}(t))
- 	\mathbb{E} [ \Dual(\boldsymbol{\dualvar}(t)) ]	
+ \textstyle \sum_{\user = 1}^{\numberofusers}	\Delta \Dual (\Delta \boldsymbol{\dualvar}_{\user}(t)) \\ \nonumber
 - \textstyle \sum_{\user = 1}^{\numberofusers}	\Delta \Dual (\Delta \boldsymbol{\dualvar}_{\user}^{\star}(t)).	\nonumber
\end{multline}
From \eqref{beta} and following definition $ \mathbb{E} [ \Dual(\boldsymbol{\dualvar}(t)) ] = \textstyle \sum_{i=0}^{\numberofusers} \Delta \Dual (0) $,

\begin{align}
& \quad \mathbb{E} [\Dual(\boldsymbol{\dualvar}(t + 1)) - \Dual(\boldsymbol{\dualvar}(t))] \nonumber \\& \geq \textstyle \sum_{\user = 1}^{\numberofusers}	\Delta \Dual (\Delta \boldsymbol{\dualvar}_{\user}^{\star}(t))
- 	\mathbb{E} [ \Dual(\boldsymbol{\dualvar}(t)) ] \nonumber 	
\\ & \hspace{3cm} - \beta \big\{ \textstyle \sum_{\user = 1}^{\numberofusers}	\Delta \Dual (\Delta \boldsymbol{\dualvar}_{\user}^{\star}(t))
-	\mathbb{E} [ \Dual(\boldsymbol{\dualvar}(t))] \big\}	\nonumber \\
&= (1-\beta) \big\{ \textstyle \sum_{\user = 1}^{\numberofusers}	\Delta \Dual (\Delta \boldsymbol{\dualvar}_{\user}^{\star}(t))
-\mathbb{E} [ \Dual(\boldsymbol{\dualvar}(t))] \big\}.																\nonumber
\end{align}
However by selecting subset of users per iteration and repeating up to only $ t $ number of communication iterations the bound is lower bounded as below,

\begin{align}
&\mathbb{E} [\Dual(\boldsymbol{\dualvar}(t + 1)) - \Dual(\boldsymbol{\dualvar}(t))] \nonumber \\
&\geq (1-\beta) \big\{ \textstyle \sum_{\user = 1}^{\numberofusers}	\Delta \Dual (\Delta \boldsymbol{\dualvar}_{\user}^{\star}(t))
- 	\mathbb{E} [ \Dual(\boldsymbol{\dualvar}(t))] \big\}																\nonumber \\
& \geq (1-\beta) \big( \textstyle \sum_{\tau=1}^{t} \textstyle \sum_{i=1}^{\numberofusers} \frac{\datasetsize_\user}{t \datasetsize} \scheduled_\user(t) \big) \big\{ \sum_{\user = 1}^{\numberofusers}	\Delta \Dual (\Delta \boldsymbol{\dualvar}_{\user}^{\star}(t))
\nonumber \\ & \hspace{6.4cm}-\mathbb{E} [ \Dual(\boldsymbol{\dualvar}(t))] \big\} \nonumber
\end{align}

Now, following \cite[Appendix B]{yang2019scheduling}, 
$ \{ \sum_{\user = 1}^{\numberofusers}	\Delta \Dual (\Delta \boldsymbol{\dualvar}_{\user}^{\star}(t))
- 	\mathbb{E} [\Dual(\boldsymbol{\dualvar}(t))] \} \geq \bar{s}\big\{\Dual (\boldsymbol{\dualvar}^{\star})
- 	\mathbb{E}[\Dual(\boldsymbol{\dualvar}(t))] \big\} $, where $ \bar{s} \in (0,1) $, we can have,
\begin{align}
 \e(\trainingDuration) &=  \mathbb{E}[\Dual(\boldsymbol{\dualvar}^\star) - \Dual(\boldsymbol{\dualvar}(T+1))] \nonumber \\
& = \mathbb{E}[\Dual(\boldsymbol{\dualvar}^\star) - \Dual(\boldsymbol{\dualvar}(T))] - \mathbb{E}[\Dual(\boldsymbol{\dualvar}(T+1)) - \Dual(\boldsymbol{\dualvar}(T))] \nonumber \\
& \leq \mathbb{E}[\Dual(\boldsymbol{\dualvar}^\star) - \Dual(\boldsymbol{\dualvar}(T))] - \nonumber \\ & \quad (1-\beta) \big( \textstyle \sum_{\tau=1}^{t}\sum_{i=1}^{\numberofusers} \frac{\datasetsize_\user}{T \datasetsize} \scheduled_\user(t) \big) \big\{\Dual (\boldsymbol{\dualvar}^{\star})
- 	\mathbb{E} [\Dual(\boldsymbol{\dualvar}(T))] \big\} \nonumber \\
&= \big( 1-(1-\beta) \textstyle \sum_{\tau=1}^{T}\sum_{i=1}^{\numberofusers} \frac{\datasetsize_\user}{T \datasetsize} \scheduled_\user(t) \big) \big\{\Dual (\boldsymbol{\dualvar}^{\star})
\nonumber \\ & \hspace{6cm}-\mathbb{E}[\Dual(\boldsymbol{\dualvar}(T))] \big\} \nonumber \\
&\leq \big( 1-(1-\beta) \textstyle\sum_{\tau=1}^{T} \sum_{i=1}^{\numberofusers} \frac{\datasetsize_\user}{T \datasetsize} \scheduled_\user(t) \big)^T \big\{\Dual (\boldsymbol{\dualvar}^{\star})
\nonumber \\ & \hspace{6cm}- \mathbb{E} [\Dual(\boldsymbol{\dualvar}(0))] \big\} \nonumber
\end{align}

In \cite{smith2016cocoa}, it is proved that $ \big\{\Dual (\boldsymbol{\dualvar}^{\star})
- 	\mathbb{E} [\Dual(\boldsymbol{\dualvar}(0))] \big\} < \datasetsize $.
Following that, 
\begin{align}
\e(\trainingDuration) &\leq  \bigg( 1-(1-\beta) \sum_{\tau=1}^{T}\sum_{i=1}^{\numberofusers} \frac{\datasetsize_\user}{T \datasetsize} \scheduled_\user(t) \bigg)^T\datasetsize. \nonumber
\end{align}

\subsection{Proof of Theorem 2}
\label{Lyapunovupperbound}
Using the inequality $ \max ( 0 , x)^2 \leq x^2 $, on \eqref{eq:queue} we have,

\begin{subequations}\label{maxinequality}
	\begin{eqnarray}
	&\frac{\q^2(t+1)}{2} \leq \frac{\q^2(t)}{2} + \frac{\big( \auxilary(t) - u(t) \big)^2}{2} + \q(t)\big( \auxilary(t) - u(t) \big),\\
	&\frac{\qX^2(t+1)}{2} \leq \frac{\qX^2(t)}{2} + \frac{\big(\auxilaryX(t) -
		\sum_{\user} \knowledgevec_\user\transpose(t) \resourceallocationvec_\user(t) \big)^2}{2} + \qX(t)\big( \auxilaryX(t) \nonumber \\ & -
	\textstyle \sum_{\user} \knowledgevec_\user\transpose(t) \resourceallocationvec_\user(t) \big).
	\end{eqnarray}
\end{subequations}

With  $ L (t) = {\big(\q(t)^2 + \qX(t)^2\big)}/{2} $, one slot drift $ \Delta L $ can be expressed as follows:
\begin{multline}
\label{Ladd}
\Delta L \leq \mathbb{E}[\q(t)\big( \auxilary(t) - u(t) \big) + \qX(t)\big( \auxilaryX(t) - \sum_{\user} \knowledgevec_\user\transpose(t) \resourceallocationvec_\user(t) \big) \\ + \big( \auxilary(t) - u(t) \big)^2/2 + \big( \auxilaryX(t) - \textstyle\sum_{\user} \knowledgevec_\user\transpose(t) \resourceallocationvec_\user(t) \big)^2/2| \q(t),\qX(t)].
\end{multline}
$L_0$ is a uniform bound on $ \big( \auxilary(t) - u(t) \big)^2/2 + \big( \auxilaryX(t) - \sum_{\user} \knowledgevec_\user\transpose(t) \resourceallocationvec_\user(t) \big)^2/2$ for all $t$, thus \eqref{Ladd} can be expressed as follows:
\begin{multline}
\Delta L \leq \mathbb{E}[\q(t)\big( \auxilary(t) - u(t) \big) + \qX(t)\big( \auxilaryX(t) - \textstyle \sum_{\user} \knowledgevec_\user\transpose(t) \resourceallocationvec_\user(t) \big) + \\ \hspace{5cm} L_0| \q(t),\qX(t)].
\end{multline}
Adding penalty term \eqref{pen}, upper bound of DPP can be expressed as,
\begin{multline}
\Delta L
-\trade \Big( 
\datasetsize\trainingDuration\big(1-\tilde{\auxilary}(t)\big)^{\trainingDuration-1} \mathbb{E}[\auxilary (t)  | \q(t)]
+ \coeffKnowledge \mathbb{E}[\auxilaryX (t)  | \qX(t)] 
\Big)
\leq \\
\textstyle  
\mathbb{E}[\q(t)\big( \auxilary(t) - u(t) \big) 
+ \qX(t)\big( \auxilaryX(t) - \sum_{\user} \knowledgevec_\user\transpose(t) \resourceallocationvec_\user(t) \big)
+ L_0 \\
-\trade \Big( 
\datasetsize\trainingDuration\big(1-\tilde{\auxilary}(t)\big)^{\trainingDuration-1} \auxilary (t) 
+ \coeffKnowledge \auxilaryX(t) \Big)
| \q(t),\qX(t)],
\end{multline}

\subsection{{Proof of Theorem 3}}
\label{ipm2}
\Copy{proofipm}{
{
Let $ \thetaw_{\user}=\frac{\q(t)(1-\beta)\datasetsize_\user }{\datasetsize} $, $ \alphaw_{\user,\resourceblock} = \qX(t)\knowledge_{\user, \resourceblock}(t) $, and $ \mathcal{B}' $ and $ \mathcal{K}' $ be the sets of allocated resource blocks and scheduled clients respectively.
}

{
Consider a scenario with non-integer solutions at optimality, i.e., $ \resourceallocation^{\star}_{\user,\resourceblock} \in (0,1] $ for $ \resourceblock \in  \mathcal{B}' $ and $\scheduled^{\star}_{\user}\in(0,1]$ for $k\in\mathcal{K}'$.
Note that for all $ \resourceblock  \in  \mathcal{B}'$, $ \alphaw_{\user,\resourceblock} = \aw $ for some $ \aw (\geq 0)$ is held ($\because$ if $\alphaw_{\user,\resourceblock'} > \aw$ for some $\resourceblock'\in \mathcal{B}'$, then optimality holds only when $ \resourceallocation_{\user,\resourceblock'} = 1$ and $ \resourceallocation_{\user,\resourceblock} = 0$ for all $\resourceblock  \notin \mathcal{B}'\setminus\{\resourceblock'\}$).
Moreover, optimality satisfies $ \one \transpose \resourceallocationvec_{\user} = 1 $.
With the constraint \eqref{eq:scheduleresource}, this yields $\sum_{\resourceblock \in \mathcal{B}}  \alphaw_{\user,\resourceblock} \resourceallocation_{\user,\resourceblock} = \sum_{\resourceblock \in \mathcal{B'}} \aw \resourceallocation_{\user,\resourceblock} = \aw$.
Hence, assigning $ \resourceallocation_{\user,\resourceblock'} = 1$ for any $\resourceblock'\in\mathcal{B}'$ with $ \resourceallocation_{\user,\resourceblock''} = 0 $ for all $\resourceblock'' \notin \mathcal{B}'\setminus\{\resourceblock'\}$ satisfies all constraints while resulting in the same optimal value, i.e., $ \one \transpose\resourceallocationvec_{\user} = \aw $. Hence, it can be noted that a solution with non-integer $ \resourceallocationvec_{\user}\optimal $ is not unique and there exists a corresponding integer solution for $ \resourceallocationvec_{\user}\optimal $  [claim A].
}

{
Substituting the above result in \eqref{eq:scheduleresource} yields $ \scheduled_{\user} \leq 1 $ and hence, \eqref{eq:scheduleresource} and \eqref{cns:relaxed_constrains} overlap.
Following the same argument as for $ \resourceallocationvec_{\user}\optimal $, it can be shown that there exists an integer solution for $ \scheduled_{\user}\optimal $ that yields the same optimal value as  with $\scheduled_{\user}\optimal \in (0,1]$ for $\user \in \mathcal{K}'$ [claim B].}

{
Based on the claims A and B, it can be noted that any non-integer optimal solution is not unique, and there exists at least one integer solution for $ \schedulingMatrix\optimal  $ and $ \resourceallocationMat\optimal$. Hence, by solving \eqref{eq:problem_per_t} using \gls{ipm} and then selecting integer values for $ \schedulingMatrix\optimal  $ and $ \resourceallocationMat\optimal $, the optimal solution of \eqref{eq:per_tbeforerelax} is obtained.}
}

\bibliographystyle{plain}
\bibliography{bibtex}

\begin{IEEEbiography}[{\includegraphics[width=1in,height=1.25in,clip,keepaspectratio]{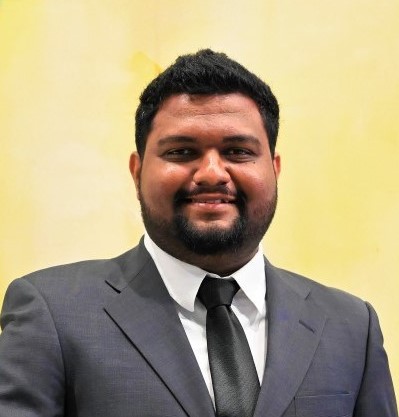}}]{Madhusanka Manimel Wadu}
	received his B.Sc. (Hons) degree in Electrical and Electronics Engineering from the University of Peradeniya, Sri Lanka, in 2015 and M.Sc. degree in wireless communication engineering from the University of Oulu, Finland by the end of 2019. He was a research assistant (part-time) at the University of Oulu, Finland from 2018 to 2019 and he has almost three years of industry exposure in Sri Lanka. He is currently a Ph.D. student at the University of Oulu, Finland. His research interests include artificial intelligence (AI), machine learning improvements in the perspective of communication, and wireless channel predictions.
\end{IEEEbiography}

\begin{IEEEbiography}[{\includegraphics[width=1in,height=1.25in,clip,keepaspectratio]{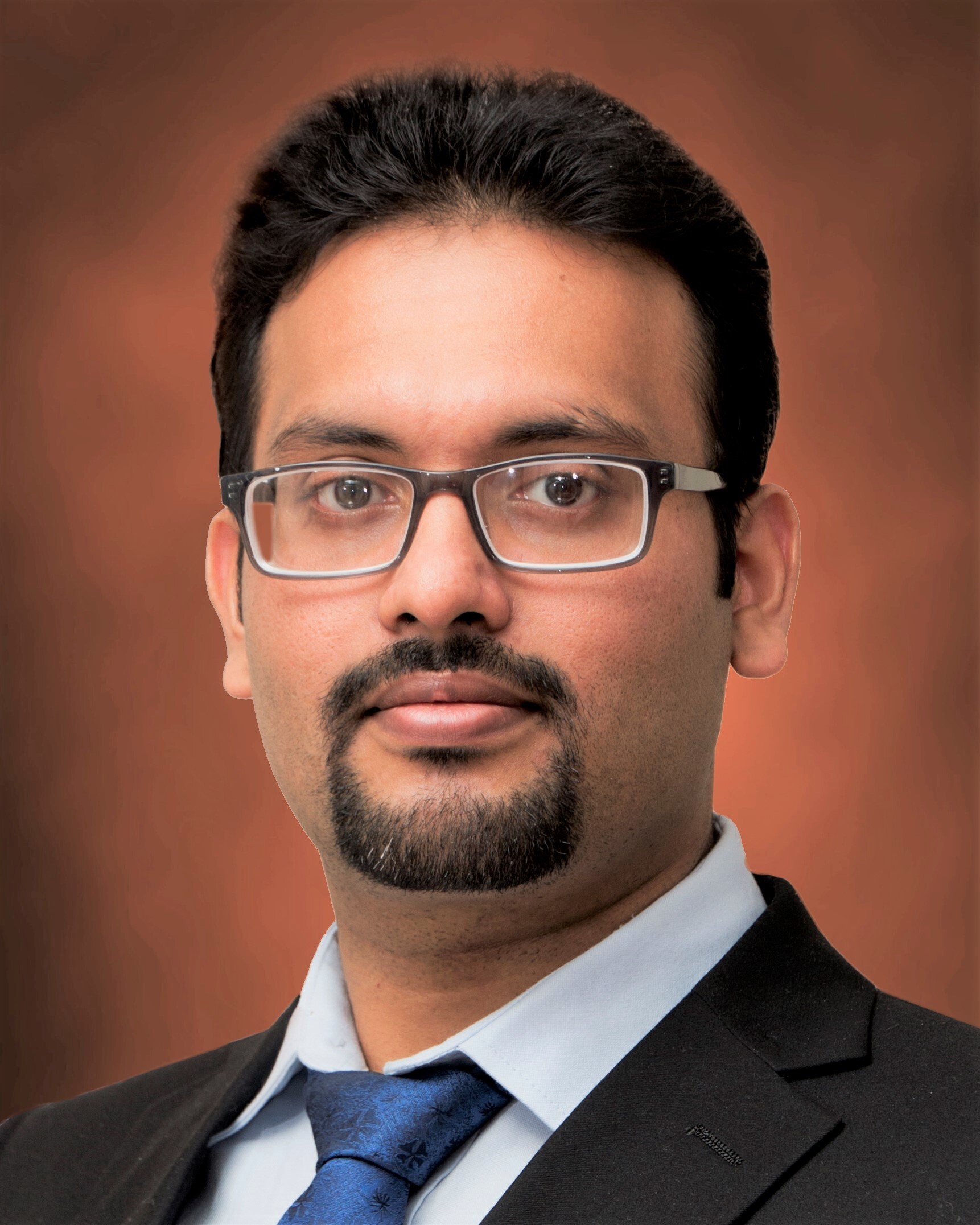}}]{Sumudu Samarakoon}
	(S'08, M'18) received the B.Sc. degree (honors) in electronic and telecommunication engineering from the University of Moratuwa, Moratuwa, Sri Lanka, in 2009, the M.Eng. degree from the Asian Institute of Technology, Khlong Nueng, Thailand, in 2011, and the Ph.D. degree in communication engineering from the University of Oulu, Oulu, Finland, in 2017. 
	He is currently a Docent (Adjunct Professor) with the Centre for Wireless Communications, University of Oulu. His main research interests are in heterogeneous networks, small cells, radio resource management, machine learning at wireless edge, and game theory. 
	Dr. Samarakoon received the Best Paper Award at the European Wireless Conference, Excellence Awards for innovators and the outstanding doctoral student in the Radio Technology Unit, CWC, University of Oulu, in 2016. 
	He is also a Guest Editor of Telecom (MDPI) special issue on ``millimeter wave communications and networking in 5G and beyond.’’
\end{IEEEbiography}
\begin{IEEEbiography}[{\includegraphics[width=1in,height=1.25in,clip,keepaspectratio]{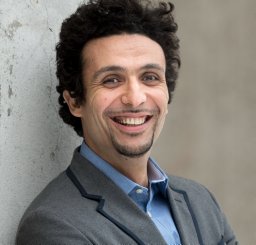}}]{Dr Mehdi Bennis}
	 is an Associate Professor at the Centre for Wireless Communications, University of Oulu, Finland, Academy of Finland Research Fellow and head of the intelligent connectivity and networks/systems group (ICON). His main research interests are in radio resource management, heterogeneous networks, game theory and distributed machine learning in 5G networks and beyond. He has published more than 200 research papers in international conferences, journals and book chapters. He has been the recipient of several prestigious awards including the 2015 Fred W. Ellersick Prize from the IEEE Communications Society, the 2016 Best Tutorial Prize from the IEEE Communications Society, the 2017 EURASIP Best paper Award for the Journal of Wireless Communications and Networks, the all-University of Oulu award for research, the 2019 IEEE ComSoc Radio Communications Committee Early Achievement Award and the 2020 Clarviate Highly Cited Researcher by the Web of Science. Dr Bennis is an editor of IEEE TCOM and Specialty Chief Editor for Data Science for Communications in the Frontiers in Communications and Networks journal. Dr Bennis is an IEEE Fellow.
\end{IEEEbiography}

\end{document}